\documentclass[conference]{IEEEtran}
\usepackage{cite}
\usepackage{amsmath,amssymb,amsfonts}
\usepackage{algorithmic}
\usepackage{graphicx}
\graphicspath{{images/}}
\usepackage{textcomp}
\usepackage{xcolor}
\usepackage{url}
\usepackage[hidelinks]{hyperref}
\def\BibTeX{{\rm B\kern-.05em{\sc i\kern-.025em b}\kern-.08em
    T\kern-.1667em\lower.7ex\hbox{E}\kern-.125emX}}
\begin{document}

\title{Learning Relative Interactions through Imitation \\ \vspace{0.5\baselineskip}
{
	\large {Università della Svizzera Italiana}\\
	{Faculty of Informatics} \\
	Lugano, Switzerland \\
	\textit{Project for the Robotics course 2019--2020}\\
}
}

\author{\IEEEauthorblockN{Giorgia Adorni}
\IEEEauthorblockA{(giorgia.adorni@usi.ch)}
\and
\IEEEauthorblockN{Elia Cereda}
\IEEEauthorblockA{(elia.cereda@usi.ch)}}

\maketitle
\thispagestyle{plain}
\pagestyle{plain}

\begin{abstract}
In this project we trained a neural network to perform specific interactions 
between a robot and objects in the environment, through imitation learning. In 
particular, we tackle the task of moving the robot to a fixed pose with respect 
to a certain object and later extend our method to handle any arbitrary pose 
around this object.

We show that a simple network, with relatively little training data, is able to 
reach very good performance on the fixed-pose task, while more work is needed 
to perform the arbitrary-pose task satisfactorily. We also explore the effect 
of ambiguities in the sensor readings, in particular caused by symmetries in 
the target object, on the behaviour of the learned controller.

\emph{External Resources}---source code~\cite{github}, pitch 
presentation~\cite{pitch} and final presentation~\cite{final-pitch}.

\end{abstract}


\section{Introduction}

In robotics, some tasks are relatively easy to perform with complete knowledge 
of the environment, but become more challenging when the environment is only 
partially observable using a robot's sensors. Imitation learning deals with 
this problem by recording the trajectories of an omniscient controller 
performing the desired task, then training a machine learning model to 
replicate them using just the data from the sensors. 

As such, the machine learning model must learn how to extract the relevant 
information from the data it receives, sidestepping the difficulty of 
implementing the perception part manually.

\begin{figure}[htbp]
	\centerline{\includegraphics[width=.8\columnwidth]{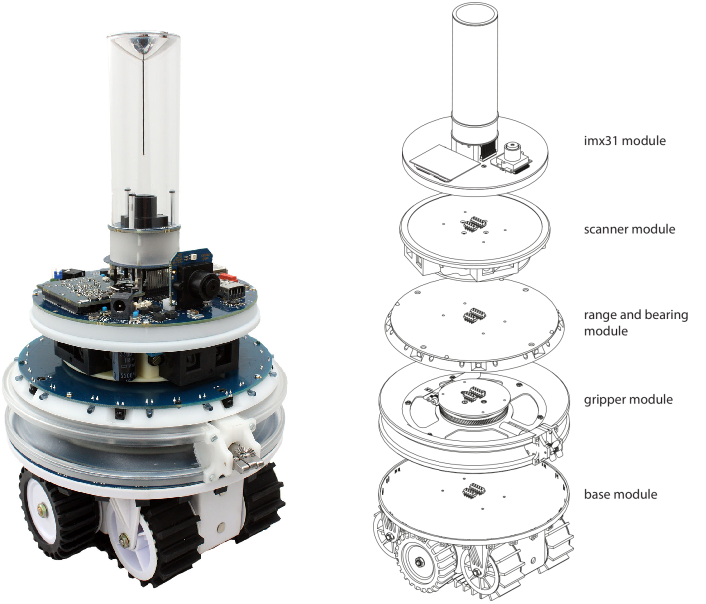}}
	\caption{Actual image and exploded CAD view of a marXbot.}
	\label{fig:marxbot}
\end{figure}

The target platform we choose for our project is the 
marXbot~\cite{bonani2010marxbot}, a research robot originally designed to study 
collective and swarm robotics. The main characteristic making the marXbot 
interesting for this project is its rotating laser scanner, which perceives 
distances and colours of the objects surrounding the robot.

The experiments are run in Enki~\cite{enki}~\cite{enki-jguzzi} a 
high-performance open-source simulator for planar robots, which provides 
collision detection and limited physics support for robots evolving on a flat 
surface. Moreover, it can simulate groups of robots hundreds of times faster 
than real-time.

We explore the symbolic task of moving the robot to a specific pose with 
respect a certain object. While easy to accomplish when the exact poses of the 
robot and the goal object are known, it becomes non-trivial when only sensor 
data is available. In particular, the model must learn to extract 
characteristic features of the object surface that allow it to determine its 
pose. Furthermore, in our case the task is complicated by inherent ambiguities 
due to the symmetry of the chosen object. Sections~\ref{sec:controllers} and 
\ref{sec:task1} explore these aspects, how we dealt with them and the results 
we obtained from the experiments.

In Section~\ref{sec:task2} we instead attempt to generalise our model, so that 
the goal pose that it should reach can be arbitrary and provided as input.

\section{Controllers}
\label{sec:controllers}

In an imitation learning setting, there are two controllers involved: an \emph{omniscient} controller, which performs the desired task with perfect knowledge of the environment, and a \emph{learned} controller, which is trained to imitate the behaviour of the omniscient controller.

\subsection{Omniscient controller}

We implemented a controller from the literature that simultaneously controls position and orientation of the robot toward a certain goal pose~\cite{park2011smooth}.

We chose this particular controller because it promised smooth and intuitive 
trajectories, which globally converge to arbitrary goal poses without 
singularities, from any initial pose. Furthermore, this specific formulation 
makes it easy to impose limits on the velocity, acceleration and jerk of the 
resulting paths, ensuring that they would be physically realizable if executed
on a real robot.

\begin{figure}[htbp]
	\centerline{\includegraphics[width=\columnwidth]{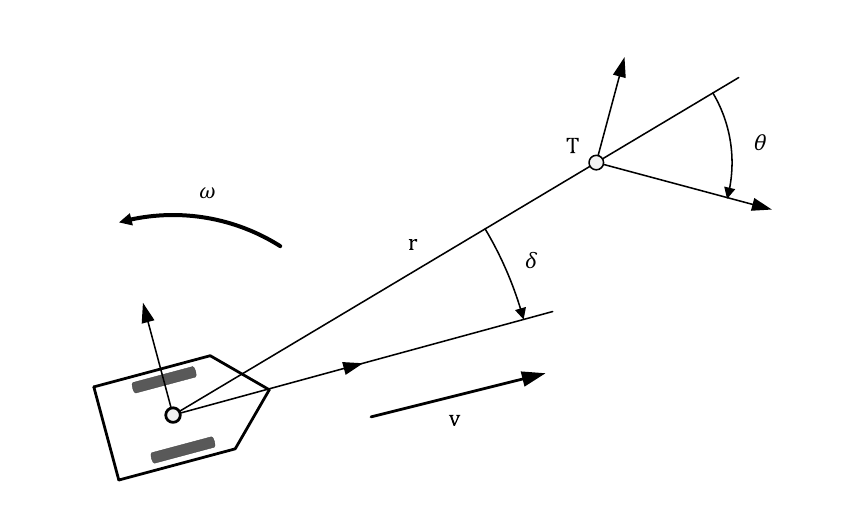}}
	\caption{Egocentric polar coordinate system, relative to the current pose of the robot.}
	\label{fig:egocentric-coordinates}
\end{figure}

The control law is described in an egocentric polar coordinate system, relative to the current pose of the robot. The control variables are the linear $v$ and angular $\omega$ velocities. Given a target pose $T$, the state of the robot is expressed as the triple $(r, \theta, \delta)$, where $r$ is the Euclidean distance from the target position; $\theta \in (-\pi, \pi]$ the target orientation with respect to the line of sight from the robot to the target position; $\delta \in (-\pi, \pi]$ the vehicle orientation with respect to the line of sight, as shown in Figure~\ref{fig:egocentric-coordinates}.

It can be seen that $(r, \theta)$ completely identify the position of the robot, while $\delta$ identifies its orientation. In this formulation, moving the robot to the target pose corresponds to bringing the state to the origin, $(r, \theta, \delta) = (0, 0, 0)$.

Assuming initially the linear velocity $v$ is nonzero positive and given (although not constant), the authors show that the angular velocity $\omega$ only influences $\delta$ directly, which in turn influences $(r, \theta)$. As such, the control problem is decomposed in a slow and a fast subsystem.

The slow subsystem first computes a reference orientation $\hat{\delta}$ to steer the robot toward the origin:
\begin{IEEEeqnarray}{l}
	\hat{\delta} = \arctan(-k_1 \theta)
\end{IEEEeqnarray}

The fast subsystem then controls the angular velocity $\omega$ to bring the current orientation $\delta$ toward the reference orientation $\hat{\delta}$ computed by the slow subsystem (somewhat confusingly, the original paper uses $\delta$ for both the current and reference orientations):
\begin{IEEEeqnarray}{l}
	\omega = -\frac{v}{r} [
		k_2 (\delta - \underbrace{\arctan(-k_1 \theta)}_{\hat{\delta}}) +
		(1 + \underbrace{\frac{k_1}{1 + (k_1\theta)^2}}_{-\dot{\hat{\delta}}})\sin\theta
	] \notag \\
	\label{eq:angular-vel}
\end{IEEEeqnarray}

It can be seen from \eqref{eq:angular-vel} that there is a linear relation between $v$ and $\omega$. In particular, $\, \omega = \kappa(r, \theta, \delta) \, v$ where $\kappa$ is the curvature of the resulting path. It is possible to rewrite \eqref{eq:angular-vel}, such that
\begin{IEEEeqnarray}{l}
	\kappa = -\frac{1}{r} [k_2 (\delta - \hat{\delta}) + (1 - \dot{\hat{\delta}})\sin\theta]
	\label{eq:curvature}
\end{IEEEeqnarray}
which implies that the shape of the path does not depend on the choice of $v$. To ensure a smooth and comfortable trajectory, the authors suggest to choose $v$ so that $v \rightarrow 0$ as $\kappa \rightarrow \pm\infty$ and $v \rightarrow v_\text{max}$ as $\kappa \rightarrow 0$:
\begin{IEEEeqnarray}{l}
	v = \frac{v_\text{max}}{1 + \beta |\kappa(r, \theta, \delta)|^\lambda}
	\label{eq:linear-vel}
\end{IEEEeqnarray}

As written, the control law has a singularity as $r \rightarrow 0$, in other words when the robot approaches the target. We address this problem as suggested in the original paper, by setting $v = k_3 r$ in the neighbourhood of $r = 0$:
\begin{IEEEeqnarray}{l}
	v' = \min(v, k_3 r)
	\label{eq:final-linear-vel}
\end{IEEEeqnarray}

\begin{figure}[htbp]
	\centerline{\includegraphics[width=.8\columnwidth]{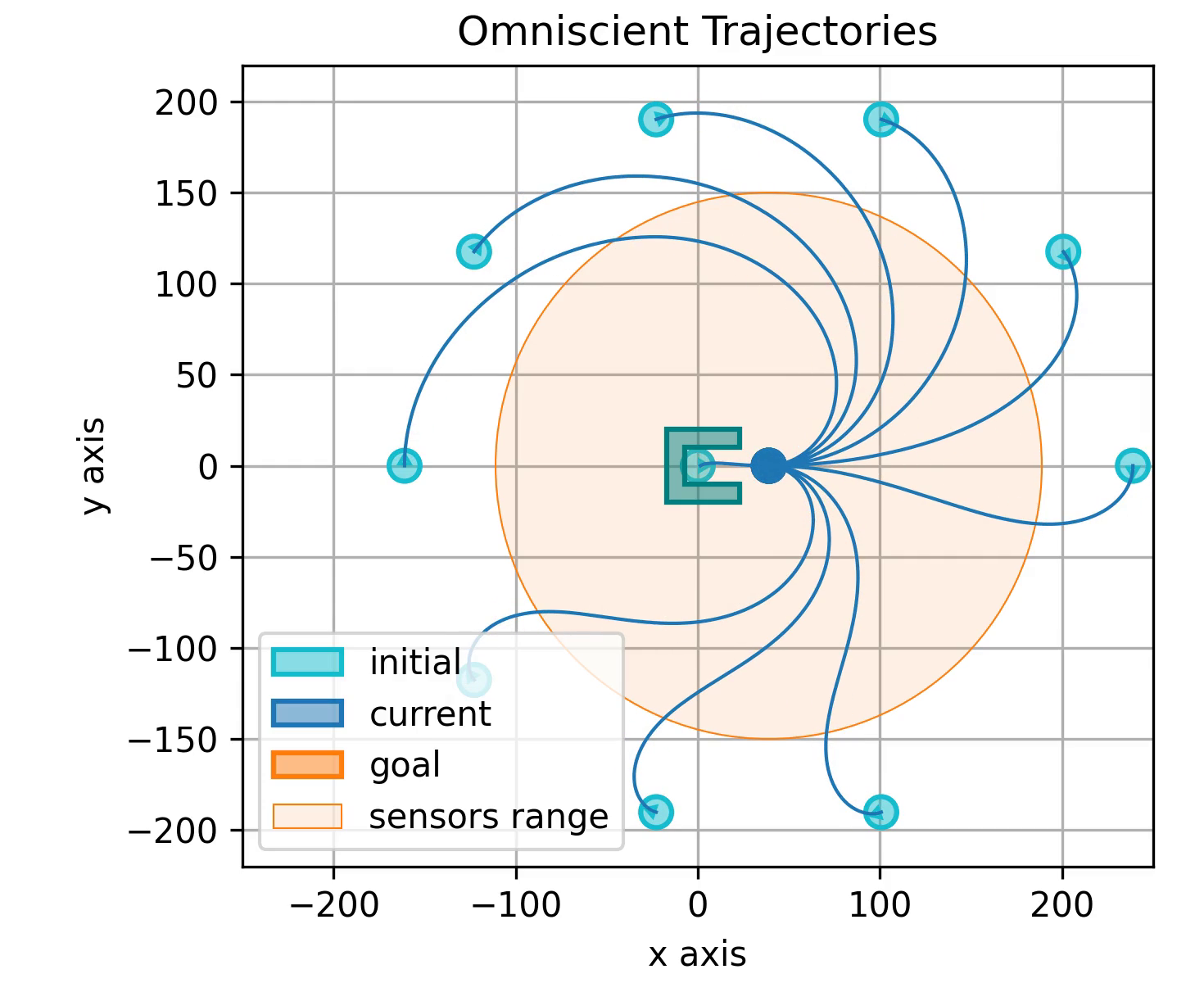}}
	\caption{Trajectories of the omniscient controller from 9 initial poses.}
	\label{fig:omniscient-trajectories}
\end{figure}

In the equations above $k_1 > 0$, $k_2 > 0$, $k_3 > 0$, $\beta > 0$ and $\lambda > 1$ are design parameters. Figure~\ref{fig:omniscient-trajectories} shows some trajectories generated by this controller, obtained with parameters $k_1 = 1$, $k_2 = 3$, $k_3 = 2$, $\beta = 0.4$ and $\lambda = 2$.

Finally, we implemented support for reverse gear as suggested by the authors: by simultaneously flipping the orientation of both the robot and the target when $v$ is negative. 

We use the reverse gear to teach the learned controller how to behave when it overshoots the target. This normally never happened with the omniscient controller, so the neural network wouldn't know what to do if it didn't stop precisely over the goal. We used an augmentation technique to ensure that this situation would be present in the training set: we made the robot move in reverse if its initial position was inside a region in front of the goal (i.e. between the arms of the object in Figure~\ref{fig:omniscient-trajectories}).

\subsection{Learned controller}

Our learned controller is implemented as an end-to-end neural network, which 
receives the sensor inputs and produces commands for the motors. 
Section~\ref{sec:models} will describe the specific architectures that we used.

\section{Task 1}
\label{sec:task1}
\subsection{Data Generation}
Relying on Enki and on the omniscient controller, a dataset of 2000 simulation 
runs is generated. 
Each run sets up a world containing:
\begin{itemize}
	\item a \emph{horseshoe-shaped object}, that represents a hypothetical 
	docking 
	station, always at pose ($x=0$, $y=0$, $\theta=0$);
	\item a \emph{fixed goal pose}, in front of the two arms of the object;
	\item a \emph{marXbot}, at a random uniform position and orientation 
	around the 
	goal, up to a maximum distance of 2 m --- slightly more that the range 
	of the distance sensors.
\end{itemize}


A simulation run is stopped either 1 s (10 time steps) after the robot 
reaches the target pose, or after 20 s (200 time steps). This ensures
enough timesteps in which the robot is standing still at the goal, improving
the performance of the network. The marXbot is considered at target if the
Euclidean distance from the goal is less than 1 mm and the robot orientation
is less than 0.5 deg from the goal orientation.

\begin{figure}[htbp]
\centerline{\includegraphics[width=\columnwidth]{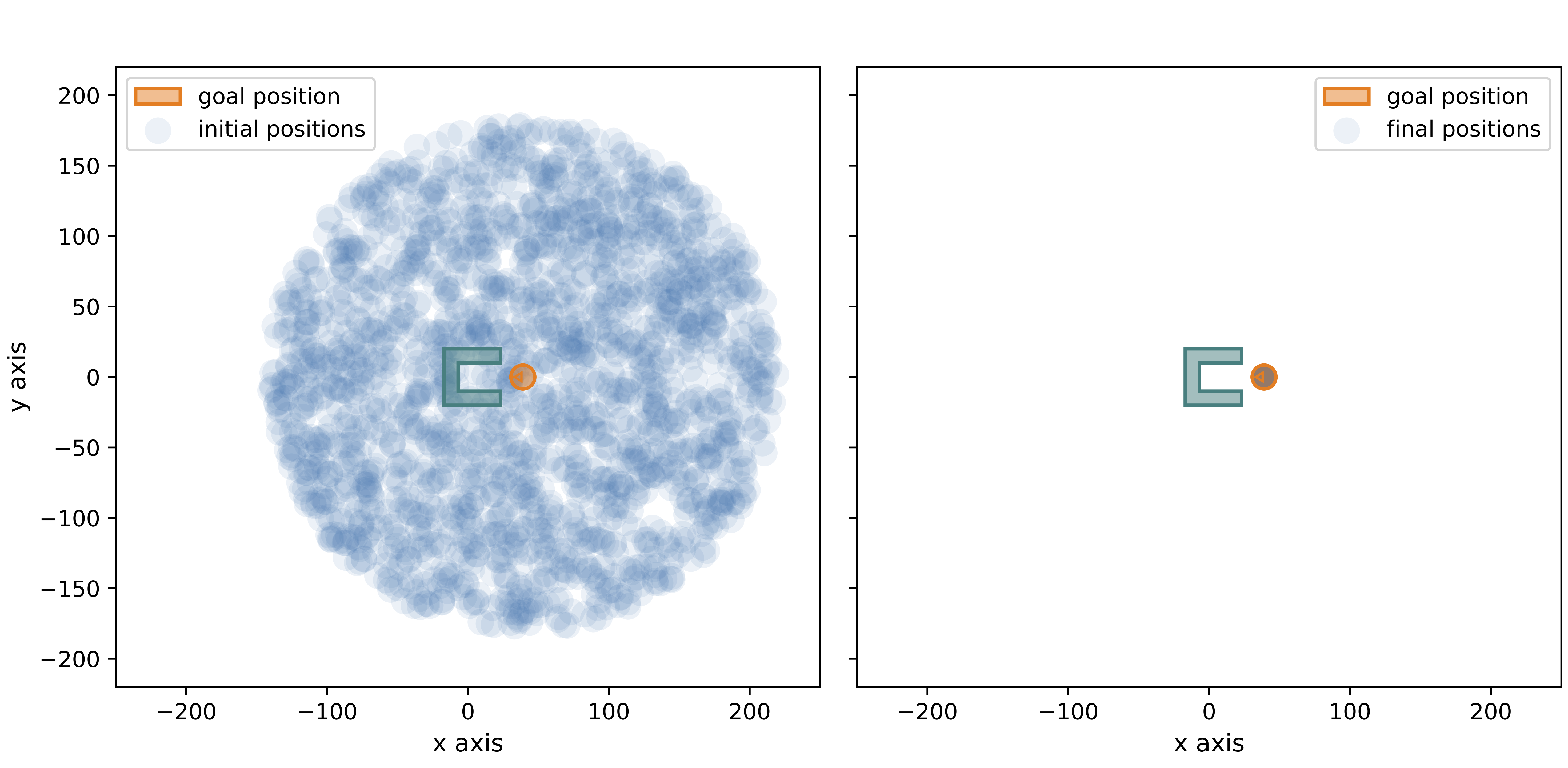}}
	\caption{Initial and final positions.}
	\label{fig:initial-final-positions-omniscient}
\end{figure}

The resulting distribution of positions at the beginning and end of each run
are shown in Figure~\ref{fig:initial-final-positions-omniscient}.

\begin{figure}[htbp]
\centerline{\includegraphics[width=.8\columnwidth]{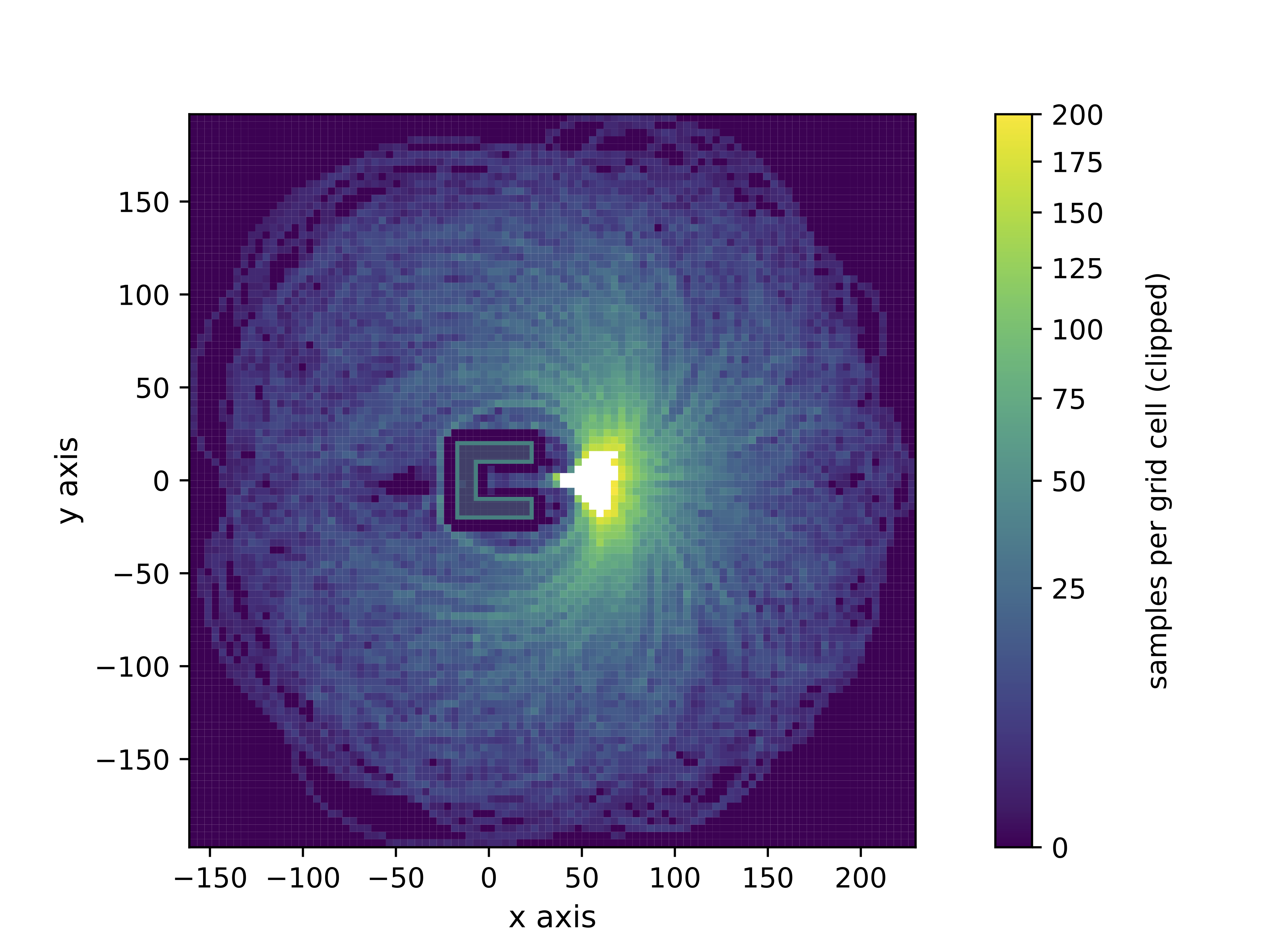}}
	\caption{Positions heat map.}
	\label{fig:densisy-omniscient}
\end{figure}

\begin{figure}[htbp]
\centerline{\includegraphics[width=.8\columnwidth]{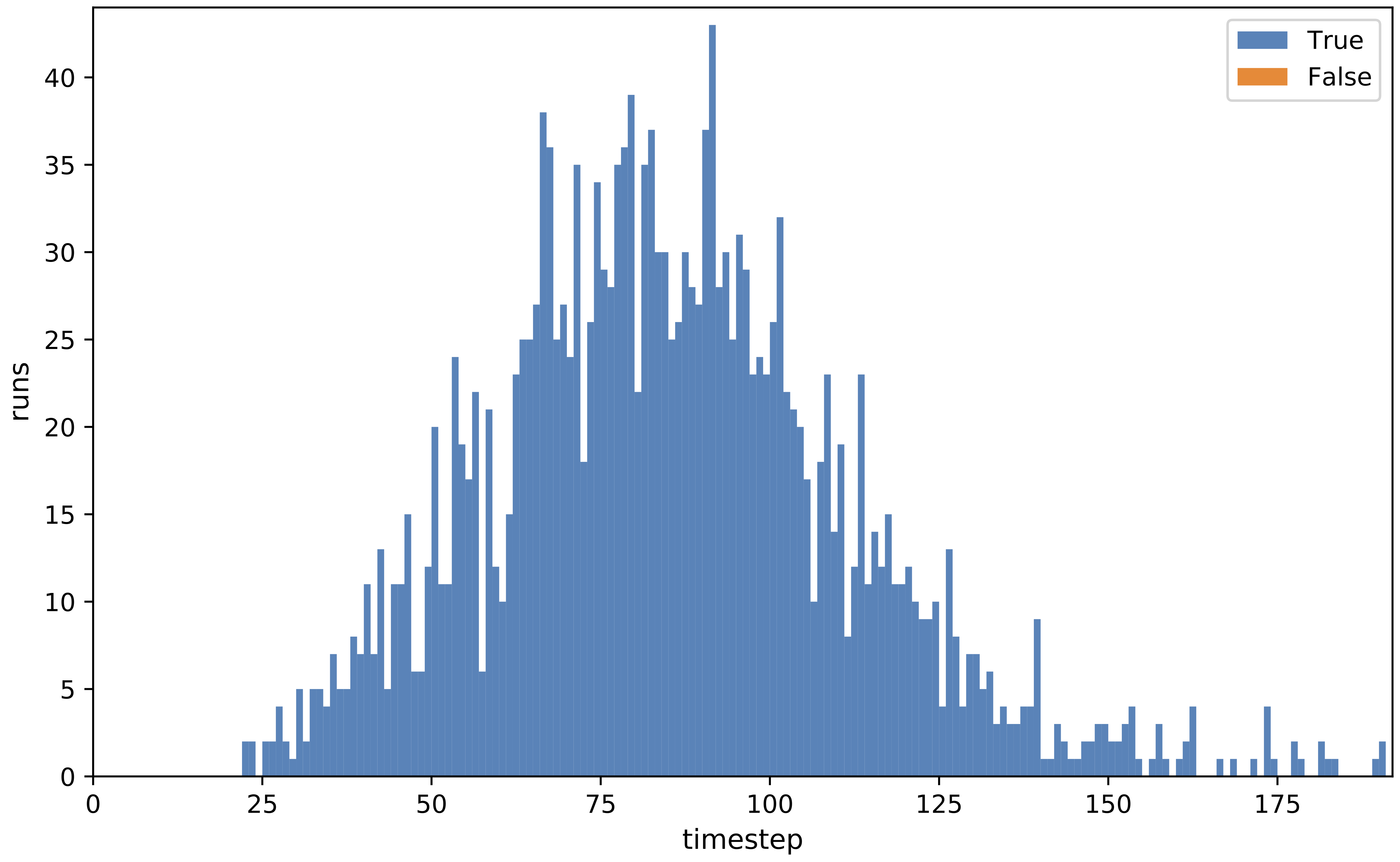}}
	\caption{Time to reach the goal.}
	\label{fig:goal-reached-omniscient}
\end{figure}

The density of samples in each position can be seen in 
Figure~\ref{fig:densisy-omniscient} while the histogram of the time needed to 
reach the goal is shown in Figure~\ref{fig:goal-reached-omniscient}. In this 
case, all the runs terminate at the goal.

\begin{figure}[htbp]
\centerline{\includegraphics[width=\columnwidth]{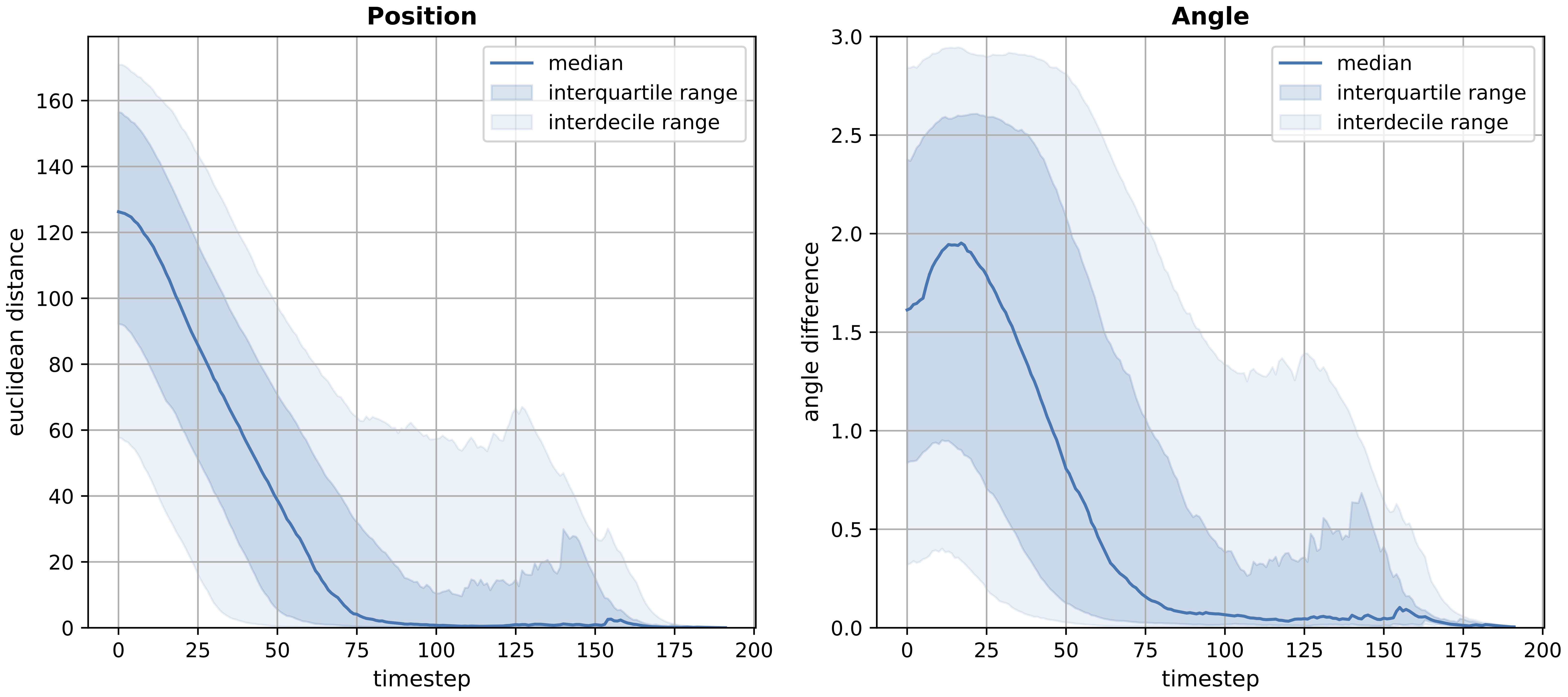}}
	\caption{Distance from goal over time.}
	\label{fig:distance-from-goal-omniscient}
\end{figure}

\begin{figure}[htbp]
\centerline{\includegraphics[width=.8\columnwidth]{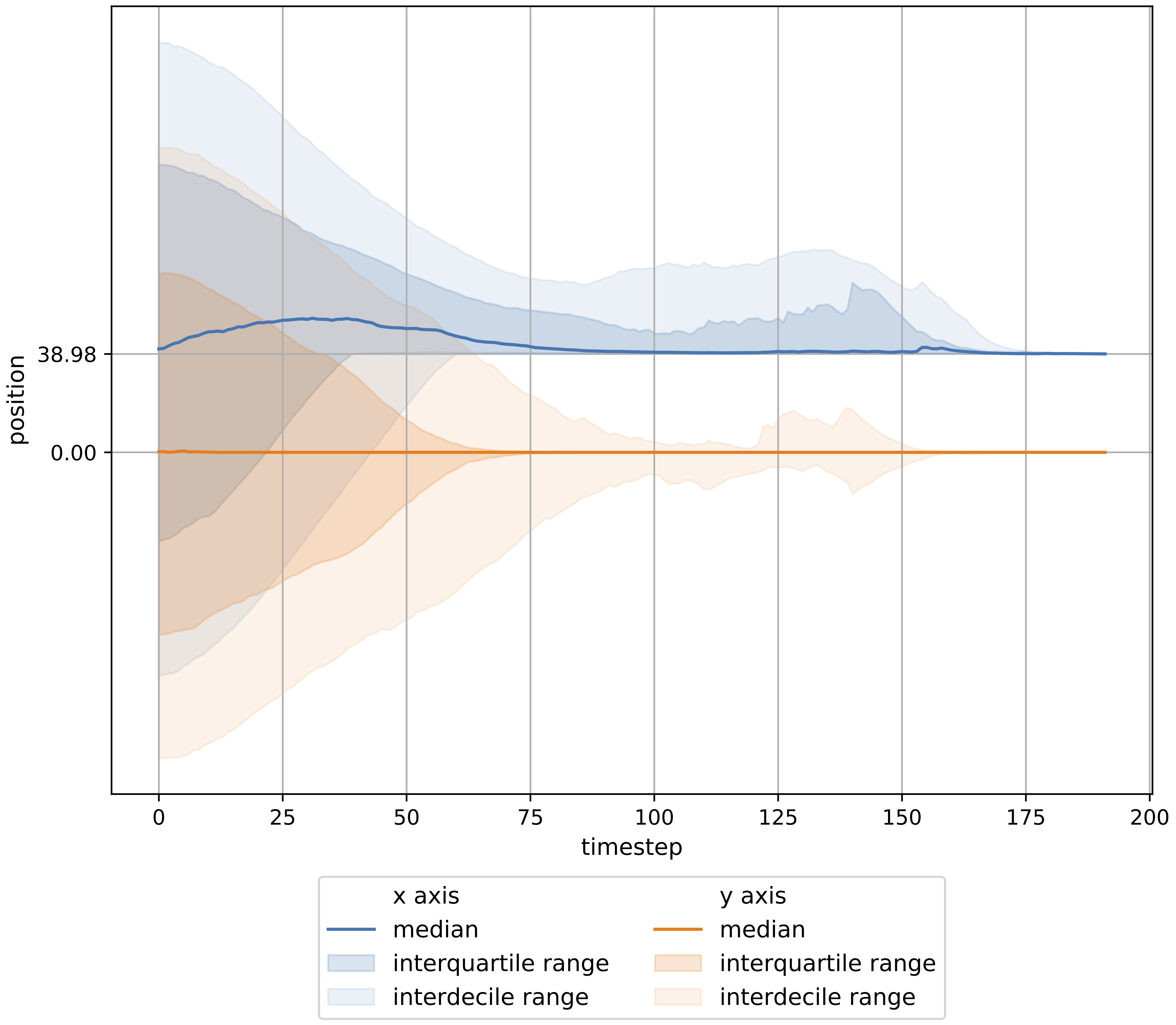}}
	\caption{Position in world coordinates over time.}
	\label{fig:position-over-time}
\end{figure}

Plotting the position and orientation errors over time shows the convergence of 
the omniscient controller. In Figure~\ref{fig:distance-from-goal-omniscient} 
are shown the Euclidean distance and the angular difference between the pose of 
the robot and the goal pose over time. Figure~\ref{fig:position-over-time} shows instead the position over time.

Finally, the generated dataset is shuffled and split, assigning each run 
into either the train (70\%, 187'000 samples), validation (15\%) or test (15\%) 
sets.

\subsection{Proposed Model}
\label{sec:models}

We designed a \emph{Convolutional Neural Network (CNN)} that takes as inputs 
the sensor distances and colour data obtained from the laser scanner 
(size: $180 \times 4$) and produces as output the left and the right wheel 
target speeds. 

One peculiarity is that we used convolutional layers with circular padding, 
since the laser scanner returns a 360° view of the world around the robot.The 
\emph{Rectified Linear Unit} (ReLU) activation function is applied after every 
layer, except for the last one. 

The input data are normalised by subtracting and dividing the channel-wise mean 
and standard deviation over the training set. Furthermore, channel-wise 
multiplicative $\alpha$ and additive $\beta$ parameters are learned during 
training to rescale the input to the most convenient range for the network:
\begin{IEEEeqnarray}{lL}
	y &= (x - \mu) / \sigma \\
	z &= \alpha y + \beta
\end{IEEEeqnarray}
This is implemented in the code with a \texttt{BatchNorm1d} layer.

The training set data are shuffled at the beginning of each epoch, so the 
mini-batches (that are of size $2^{14}$) are generated independently between 
epochs. 

The model is trained with the \emph{Adam} optimiser and learning rate $0.001$, 
while the other parameters have their default values. The training is 
interrupted using \emph{early stopping}, if the validation loss doesn't improve 
for 20 epochs, or after 500 epochs. 

During the various experiments, four different architectures are evaluated:
\begin{itemize}
	\item \emph{Baseline network}: 3 convolutional and 3 fully-connected
	layers (Table~\ref{tab: baseline})
	\item Baseline network plus one max pooling layer (Table~\ref{tab: maxpool})
	\item Baseline network plus dropout (Table~\ref{tab: baseline} + 
	Table~\ref{tab: dropout})
	\item Baseline network plus one max pooling layer and dropout 
	(Table~\ref{tab: maxpool} + Table~\ref{tab: dropout})
\end{itemize} 

\begin{table}[htbp]
	\caption{Architecture of the Baseline Network}
	\begin{center}
		\begin{tabular}{|c|c|c|c|c|}
			\hline
			\textbf{Layer}&\textbf{Channels} &\textbf{Kernel size} &\textbf{Stride} &\textbf{Padding}\\
			\cline{1-5}
			conv1 &  4 $\rightarrow$ 16 & 5 & 2 & 2, circular \\ \hline
			conv2 & 16 $\rightarrow$ 32 & 5 & 2 & 2, circular \\ \hline
			conv3 & 32 $\rightarrow$  			 32 & 5 & 1 & 2, circular \\ \hline
			fc1 &   45 $\times$ 32 $\rightarrow$ 128 &  &  &  \\ \hline
			fc2 &  128 $\rightarrow$ 128 &  &  &  \\ \hline
			fc3 &  128 $\rightarrow$   2 &  &  &  \\ \hline
		\end{tabular}
		\label{tab: baseline}
	\end{center}
\end{table}

\begin{table}[htbp]
	\caption{Architecture of the Network with Max Pooling}
	\begin{center}
		\begin{tabular}{|c|c|c|c|c|}
			\hline
			\textbf{Layer}&\textbf{Channels} &\textbf{Kernel size} &\textbf{Stride} &\textbf{Padding}\\
			\cline{1-5}
			conv1  &   4 $\rightarrow$  \bfseries	32 & 5 & 2 & 2, circular \\ \hline
			conv2  & \bfseries 32 $\rightarrow$  	96 & 5 & 2 & 2, circular \\ \hline
			\bfseries mpool1 & 					   & \bfseries 3	& \bfseries 3 & \bfseries 1, circular \\ 
			\hline			
			conv3  & \bfseries 96 $\rightarrow$  	96 & 5 & 1 & 2, circular \\ \hline
			fc1    & 15 $\times$ 96 $\rightarrow$ 128 &  &  &  \\ \hline
			fc2    & 128 $\rightarrow$ 128 &  &  &  \\ \hline
			fc3    & 128 $\rightarrow$   2 &  &  &  \\ \hline
		\end{tabular}
		\label{tab: maxpool}
	\end{center}
\end{table}

\begin{table}[htbp]
	\caption{Architecture of the Network with Dropout}
	\begin{center}
		\begin{tabular}{|c|c|c|c|c|}
			\hline
			\textbf{Layer}&\textbf{Channels} &\textbf{Kernel size} &\textbf{Stride} &\textbf{Padding}\\
			\cline{1-5}
			\multicolumn{5}{|c|}{...} \\ \hline
			fc1 &  1440 $\rightarrow$ 128 &  &  &  \\ \hline
			\bfseries drop1 & \multicolumn{4}{c|}{\bfseries dropout with p = 0.5} \\ \hline
			fc2 &  128 $\rightarrow$ 128 &  &  &  \\ \hline
			\bfseries drop2 & \multicolumn{4}{c|}{\bfseries dropout with p = 0.5} \\ \hline
			fc3 &  128 $\rightarrow$   2 &  &  &  \\ \hline
		\end{tabular}
		\label{tab: dropout}
	\end{center}
\end{table}

\subsection{Experiment 1}
\begin{figure}[htbp]
	\centerline{\includegraphics[width=\columnwidth]{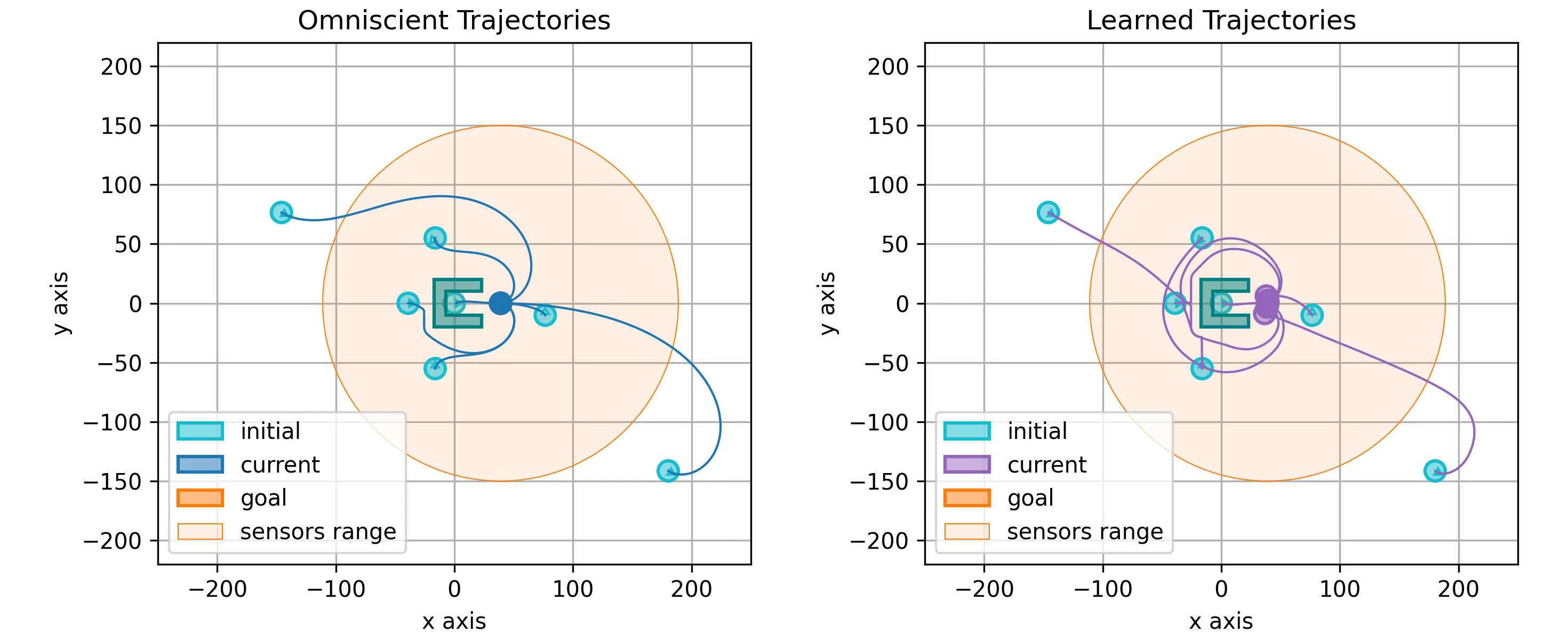}}
	\caption{Trajectories of the controller learned from the baseline network.}
	\label{fig:baseline}
\end{figure}

The first experiment performed compares the different architectures using the 
\emph{Mean Squared Error (MSE)} loss function.
The baseline model is able to reach the goal, but with little precision and 
often colliding with the object, as shown in Figure~\ref{fig:baseline}.

\begin{figure}[htbp]
	\centerline{\includegraphics[width=\columnwidth]{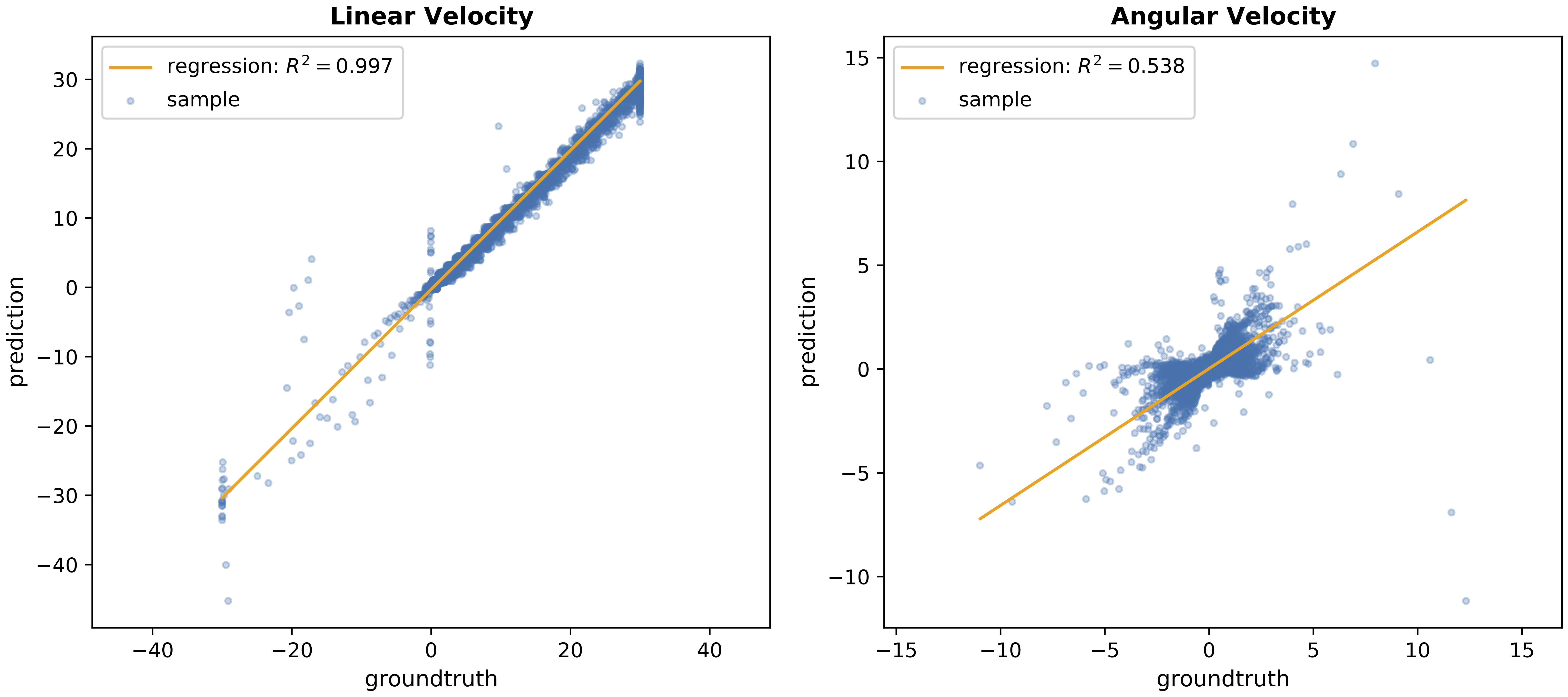}}
	\caption{$R^2$ regressor on the validation set of the baseline network.}
	\label{fig:regression-baseline}
\end{figure}

Adding either max pooling or dropout alone does not solve the problem, but 
combining them results in a visible improvement: the robot reaches the goal 
position more precisely even if oscillating a bit.

\begin{figure}[htbp]
	\centerline{\includegraphics[width=\columnwidth]{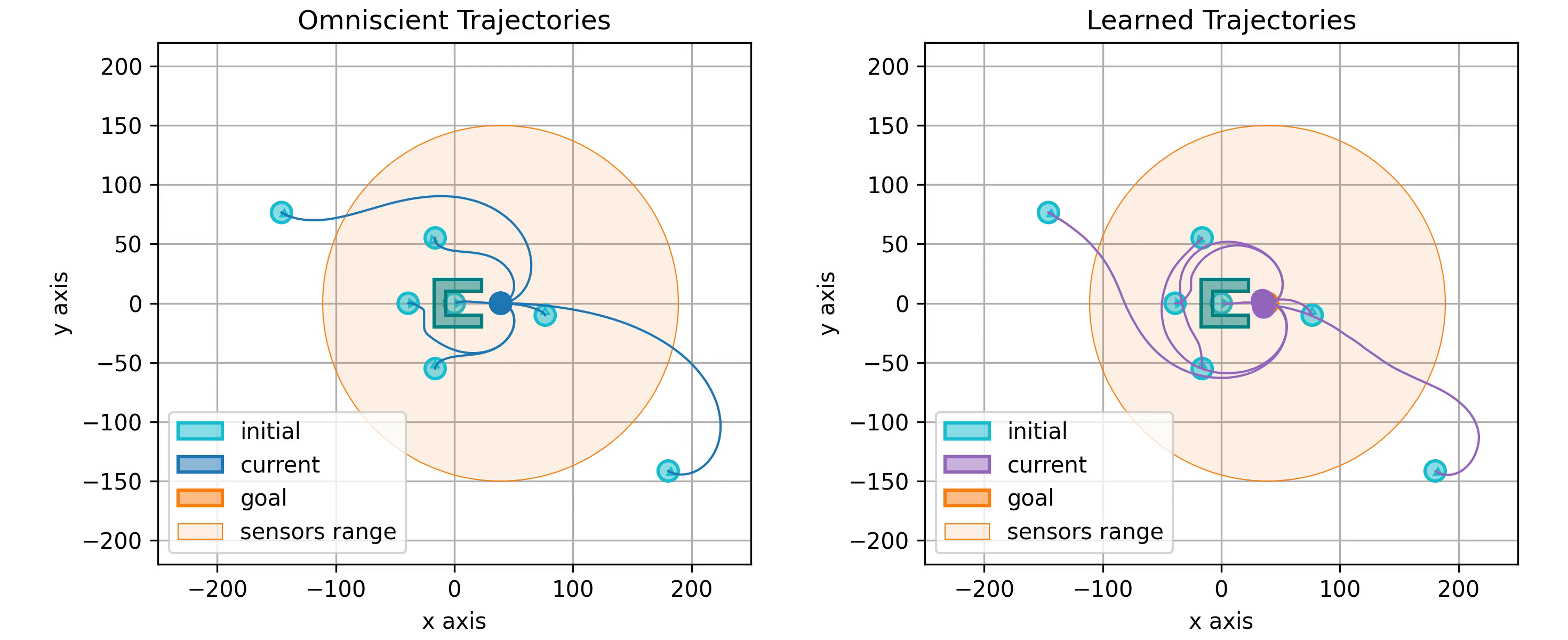}}
	\caption{Trajectories of the controller learned from the max pooling + 
	dropout network.}
	\label{fig:maxpool+dropout}
\end{figure}

\begin{figure}[htbp]
	\centerline{\includegraphics[width=\columnwidth]{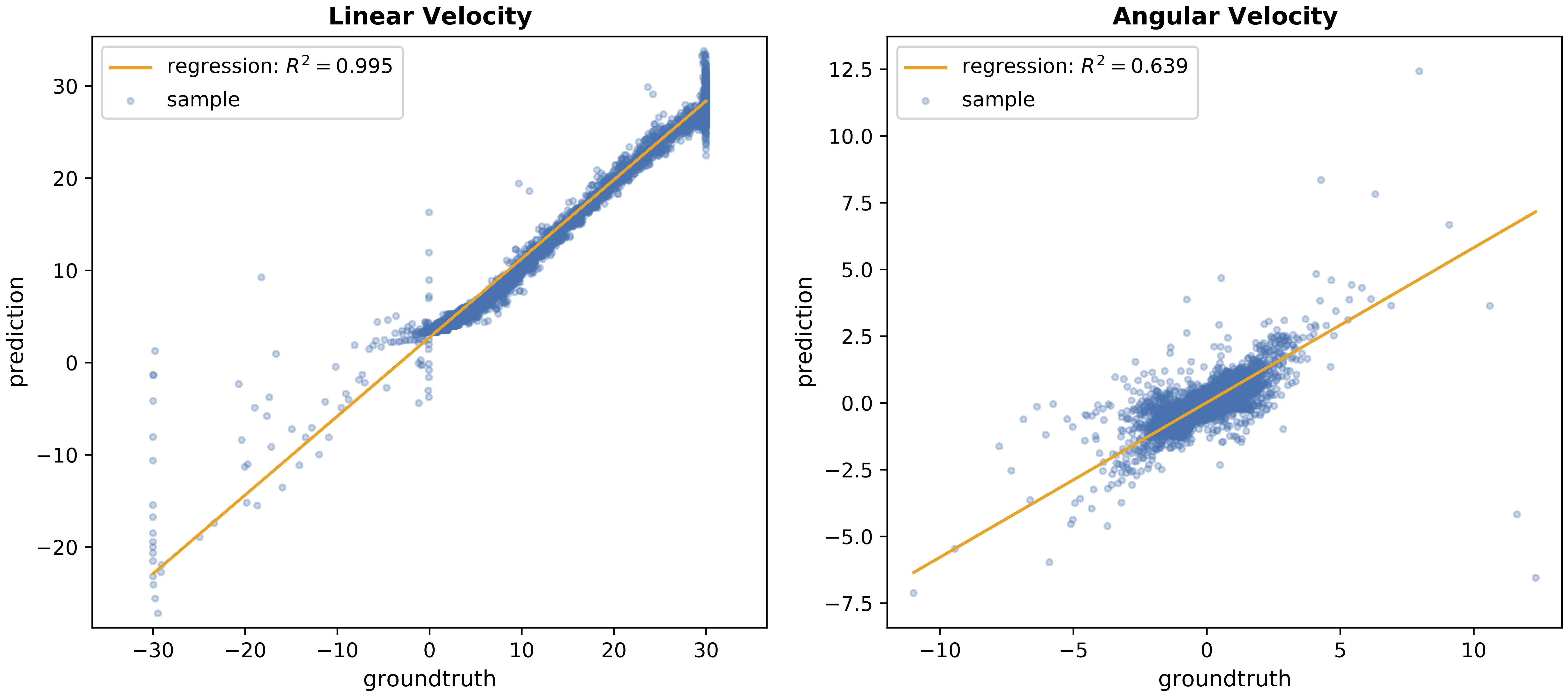}}
	\caption{$R^2$ regressor on the validation set of the max pooling + dropout 
	network.}
	\label{fig:regression-maxpool+dropout}
\end{figure}

\begin{figure}[htbp]
\centerline{\includegraphics[width=.8\columnwidth]{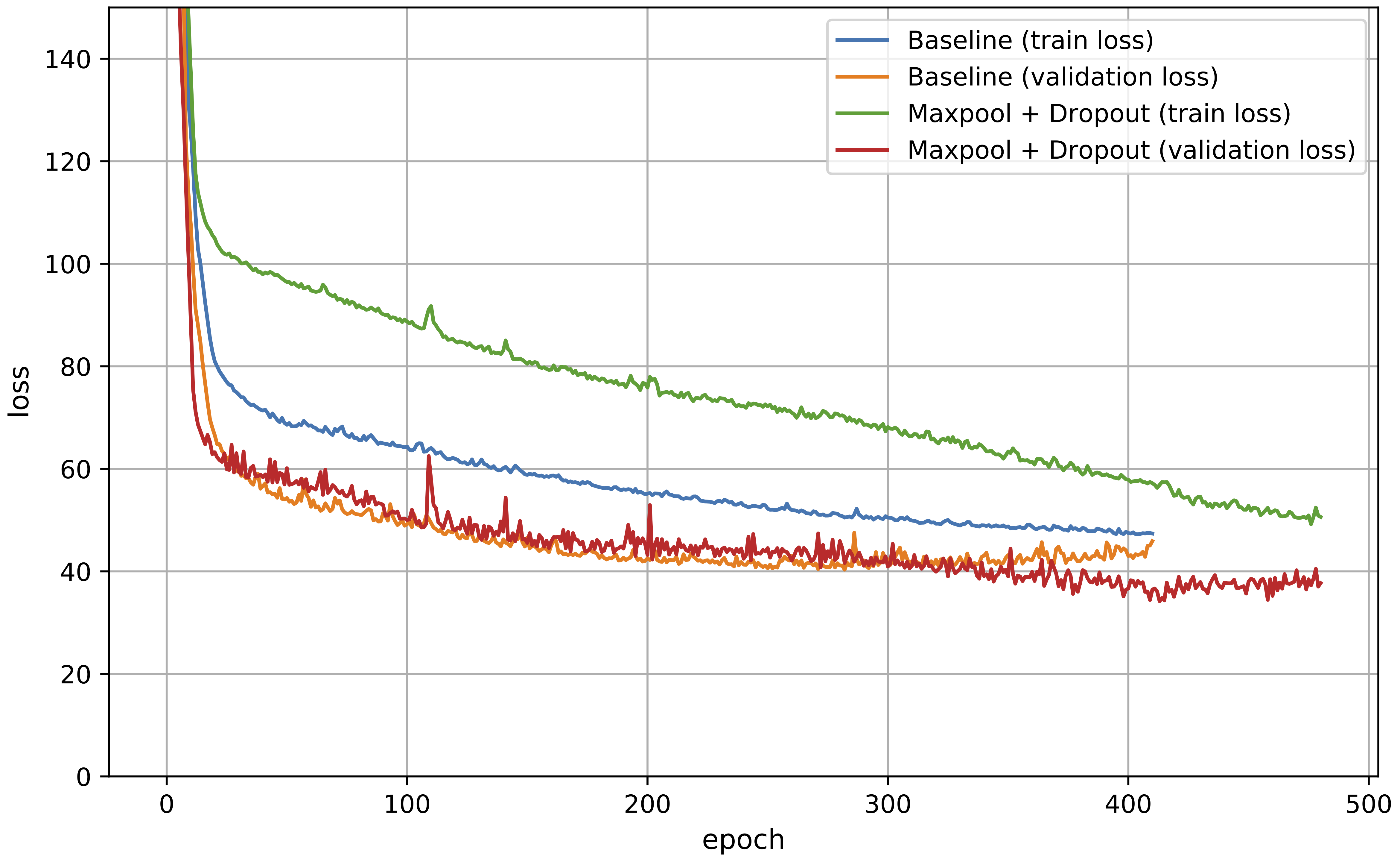}}
	\caption{Comparison of the losses among train and validation sets.}
	\label{fig:loss}
\end{figure}

The regression coefficient of the angular velocity, displayed in 
Figure~\ref{fig:regression-maxpool+dropout}, increases from $0.54$ to $0.64$ 
compared to Figure~\ref{fig:regression-baseline}, confirming the improvement of 
the second model. As shown in Figure~\ref{fig:loss}, it shows also a lower 
validation loss (in red) and does not overfit toward the end of training, like 
the baseline model (in orange).

\begin{figure}[htbp]
	\centerline{\includegraphics[width=\columnwidth]{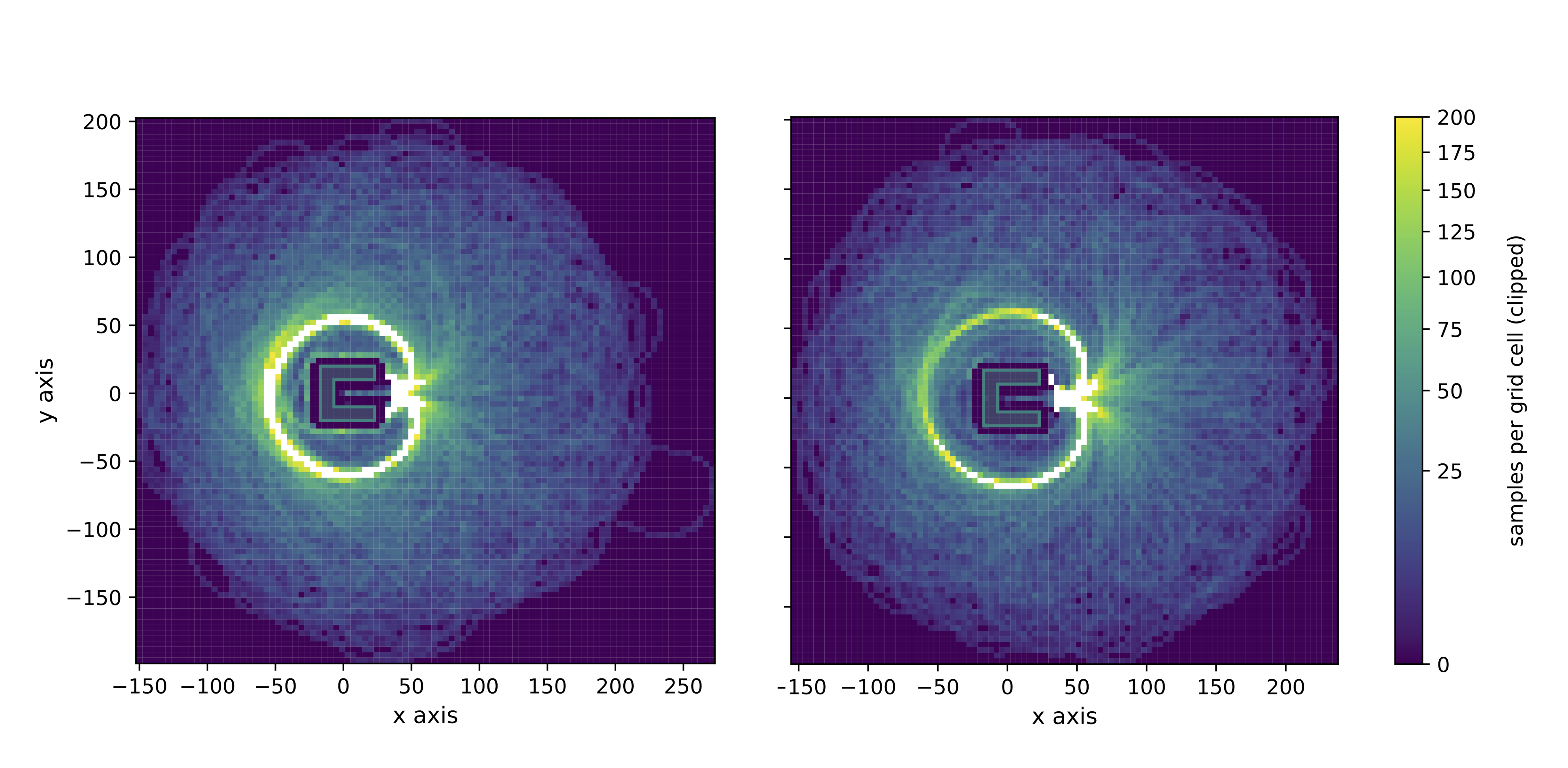}}
	\caption{Positions heat maps.}
	\label{fig:heatmaps}
\end{figure}

Finally, the end positions are more tightly clustered over the goal.
Both heat maps in Figure~\ref{fig:heatmaps} show a tendency to rotate around 
the object, which are caused by its symmetry. We will explore this issue in 
Section~\ref{experiment3}.

\begin{figure}[htbp]
	\centerline{\includegraphics[width=\columnwidth]{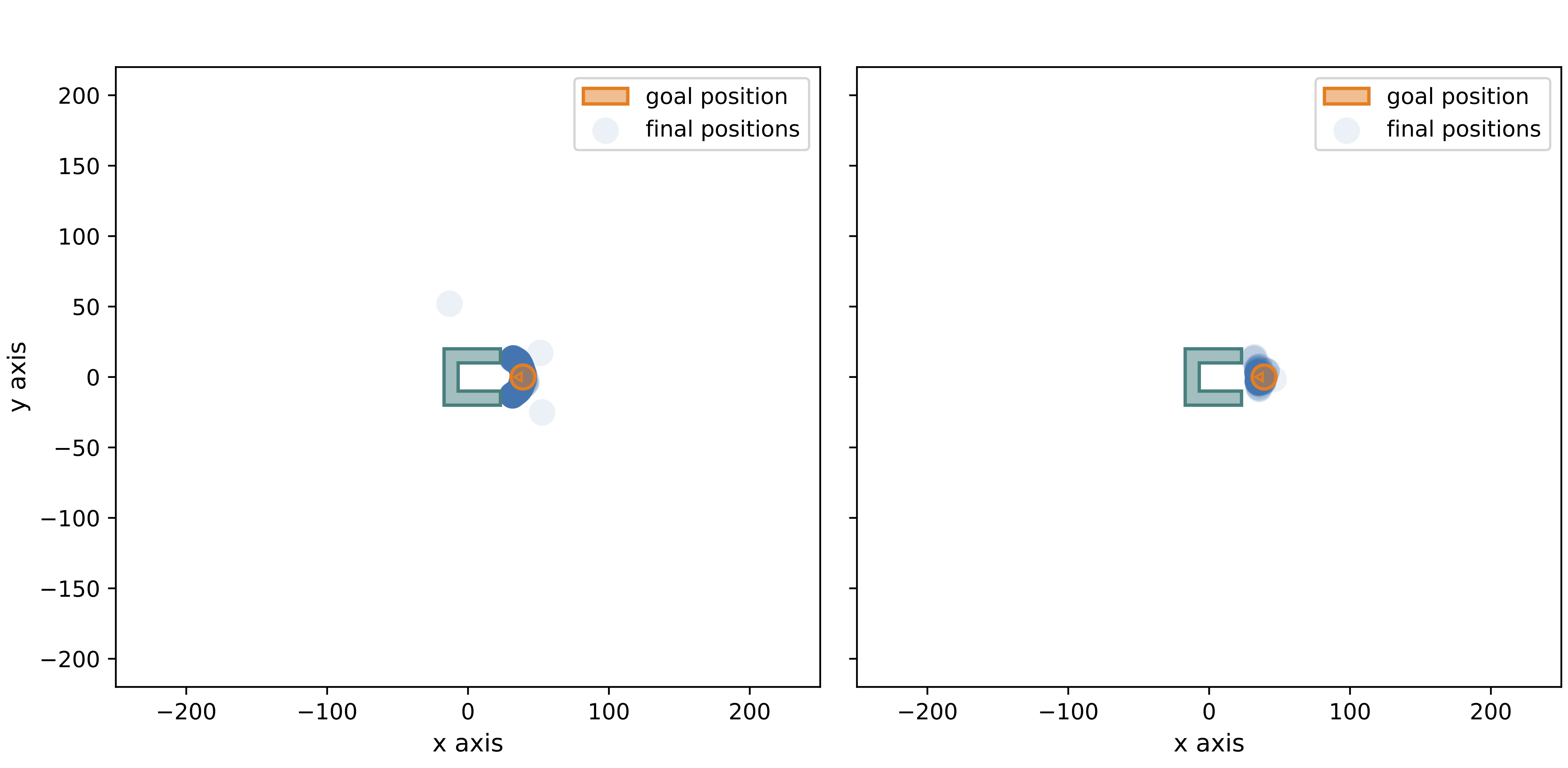}}
	\caption{Final positions.}
	\label{fig:final-positions}
\end{figure}

\begin{figure}[htbp]
	\centerline{\includegraphics[width=\columnwidth]{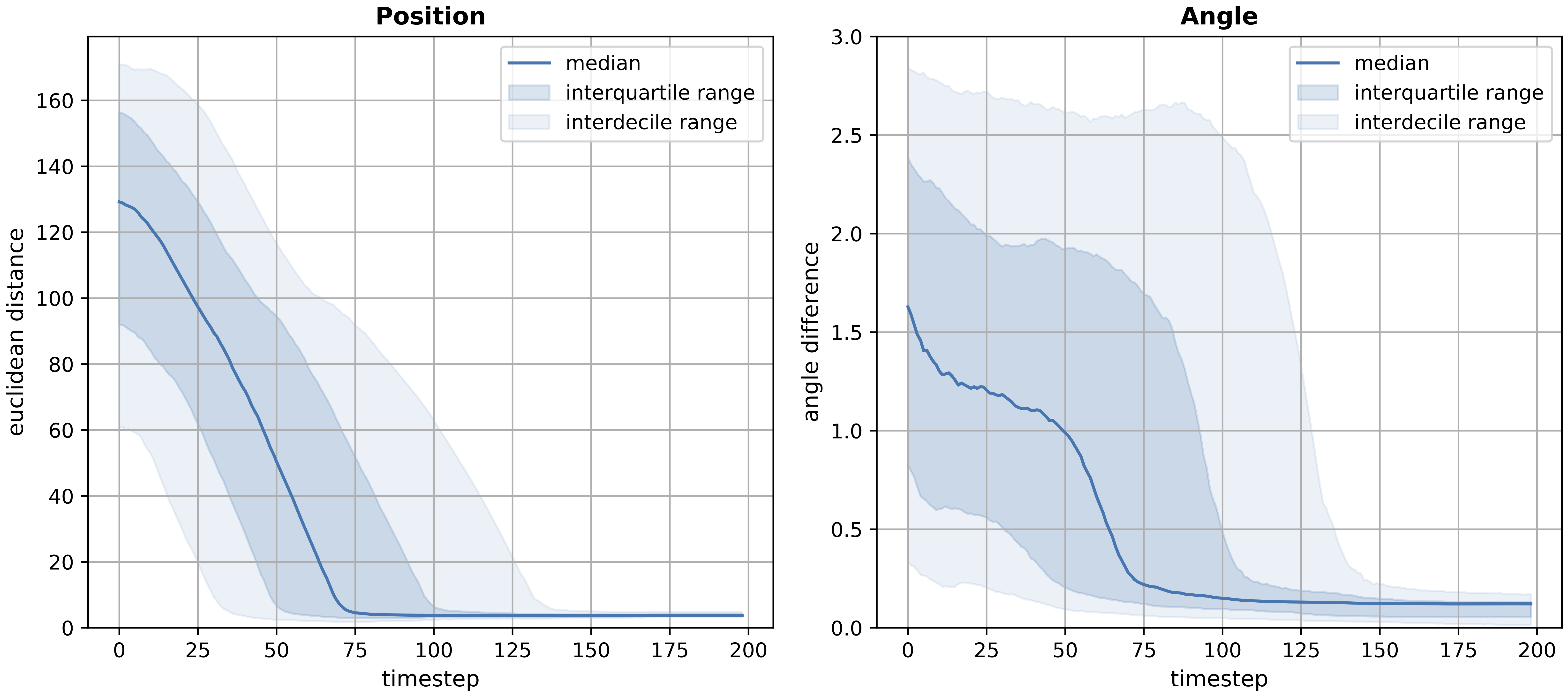}}
	\caption{Distance from goal over time.}
	\label{fig:distance-from-goal-learned}
\end{figure}

Even though the overall behaviour of this model is good, the main drawback is a 
slight systematic error in the final orientation that can be seen in 
Figure~\ref{fig:distance-from-goal-learned}.

\subsection{Experiment 2}
The second experiment evaluates the performance of the same max pooling + 
dropout model, but trained with a different loss function, \emph{Smooth 
L1} \cite{smoothl1}, which is less sensitive to outliers than \emph{MSE} and 
has been shown to prevent exploding gradients in some cases. It is computed as

\begin{equation}
\text{L}(x, y) = \frac{1}{n}\sum_{i}z_i
\label{smoothl1}
\end{equation}

where $z_i$ is given by

\begin{equation}
z_i = 
\begin{cases}
0.5 (x_i-y_i)^2, &\text{ if } |x_i-y_i|<1 \\
|x_i-y_i| - 0.5, &\text{ otherwise}
\end{cases}
\end{equation}

\begin{figure}[htbp]
	\centerline{\includegraphics[width=\columnwidth]{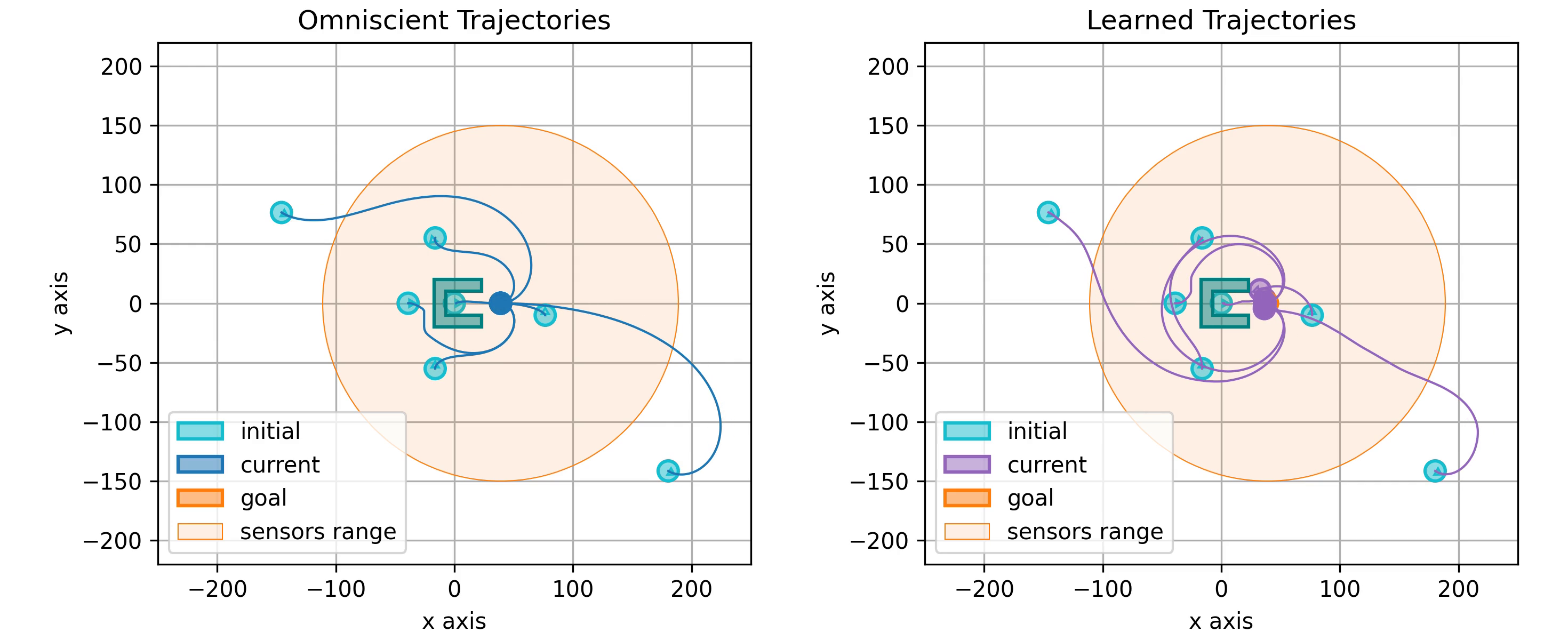}}
	\caption{Trajectories.}
	\label{fig:demo-trajectories}
\end{figure}

Although it results in less precise final positions, it solves the oscillation 
issue, as shown in Figure~\ref{fig:demo-trajectories}.

\begin{figure}[htbp]
	\centerline{\includegraphics[width=\columnwidth]{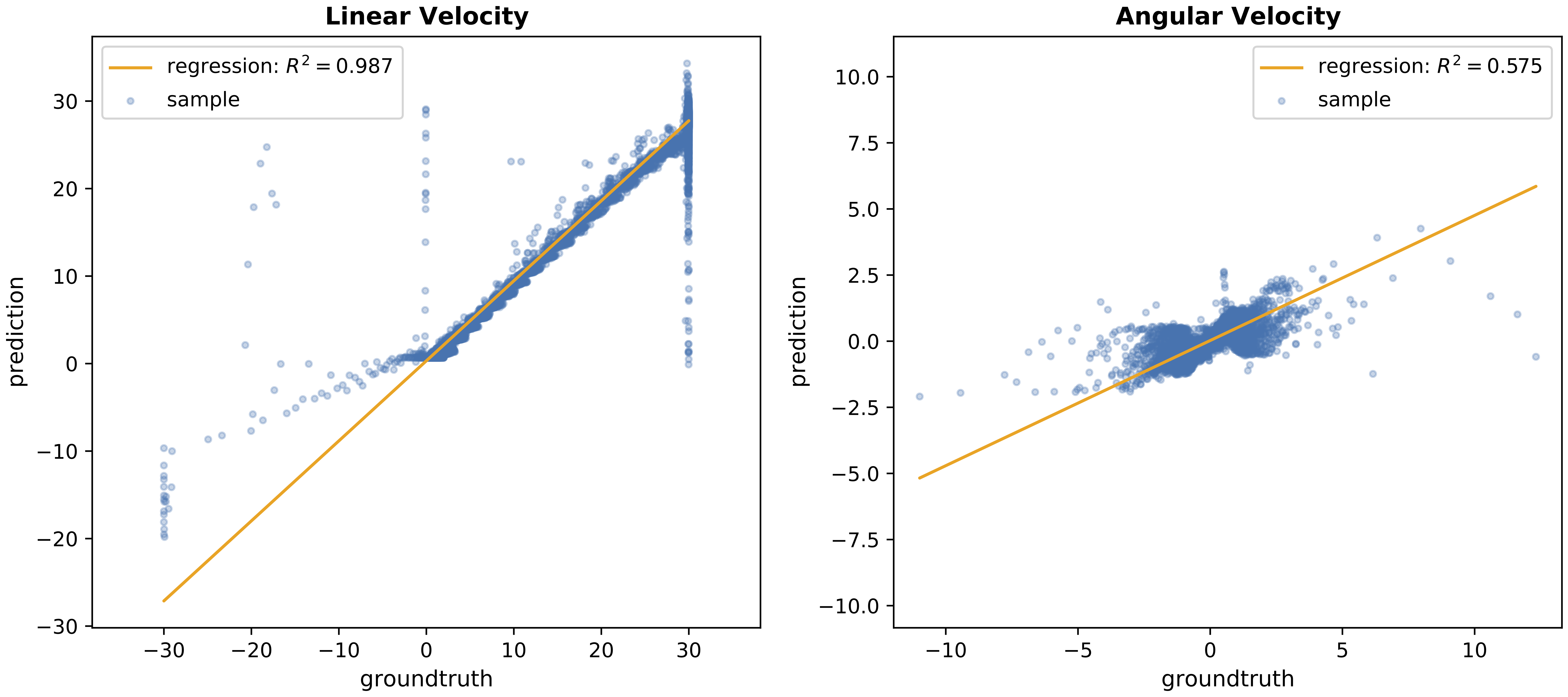}}
	\caption{$R^2$ regressor on the validation set.}
	\label{fig:regression-validation}
\end{figure}

The regression coefficient of the angular velocity, shown in 
Figure~\ref{fig:regression-validation}, decreases from $0.64$ to $0.58$, 
confirming the superiority of the previous model.

\subsection{Experiment 3}
\label{experiment3}

\begin{figure}[htbp]
	\centerline{\includegraphics[width=\columnwidth]{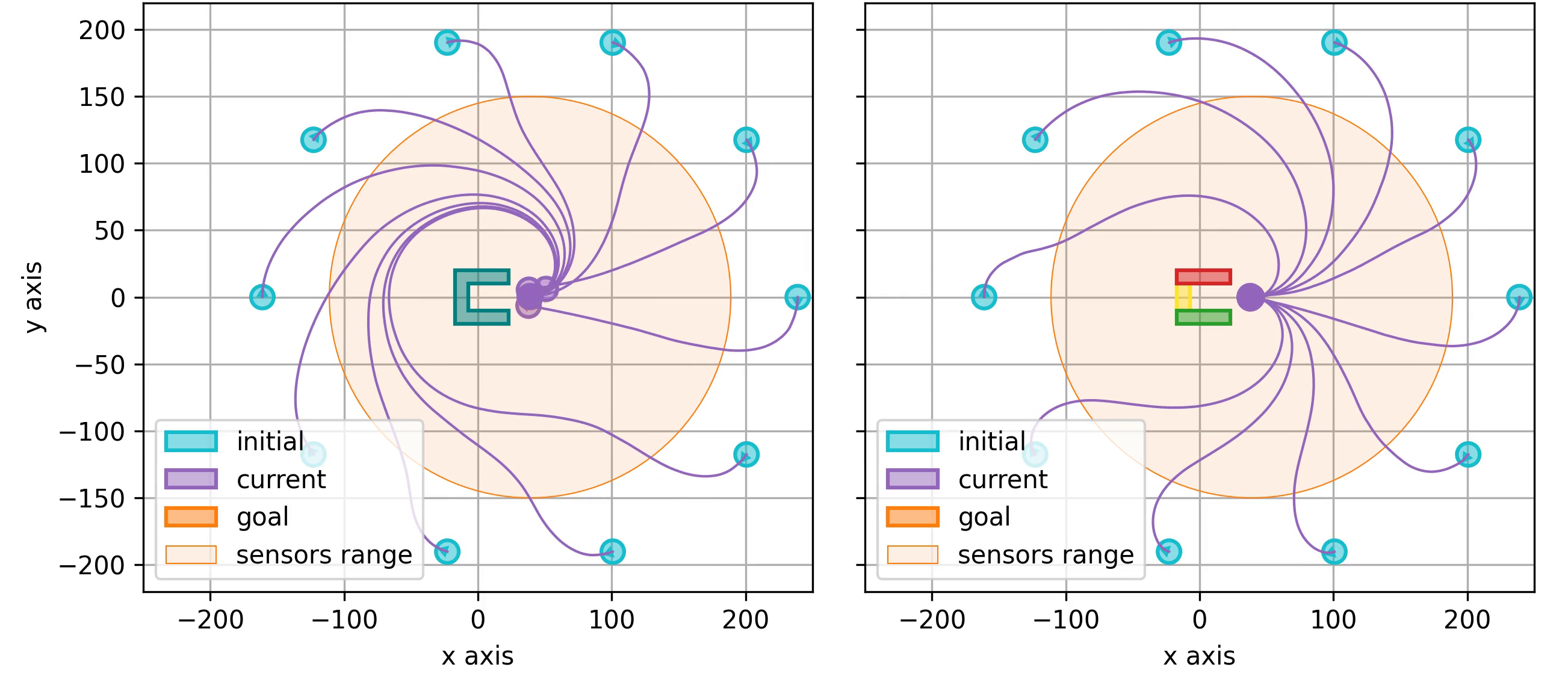}}
	\caption{Comparison of the trajectories of the monochromatic and 
	polychromatic goal object.}
	\label{fig:demo-circle-trajectories}
\end{figure}

The monochromatic goal object shown so far has symmetries that make the 
trajectory to follow ambiguous, causing the robots converge to the goal in 
sub-optimal paths. One such path is visualised in 
Figure~\ref{fig:demo-circle-trajectories}, in other cases we saw the learned 
controller always moving counter-clockwise or alternating depending on the 
initial pose. 

In either case, the controller learns a good behaviour for the available data, 
ultimately reaching the goal pose, albeit following a different path. This is 
particularly interesting, considered that we only train the network to imitate 
trajectories, with no indication of the goal.

The symmetries are addressed in this final experiment, by using a polychromatic 
goal object that has a different colour for each of its faces. This removes any 
localisation ambiguity in the sensor readings.

\begin{figure}[htbp]
	\centerline{\includegraphics[width=\columnwidth]{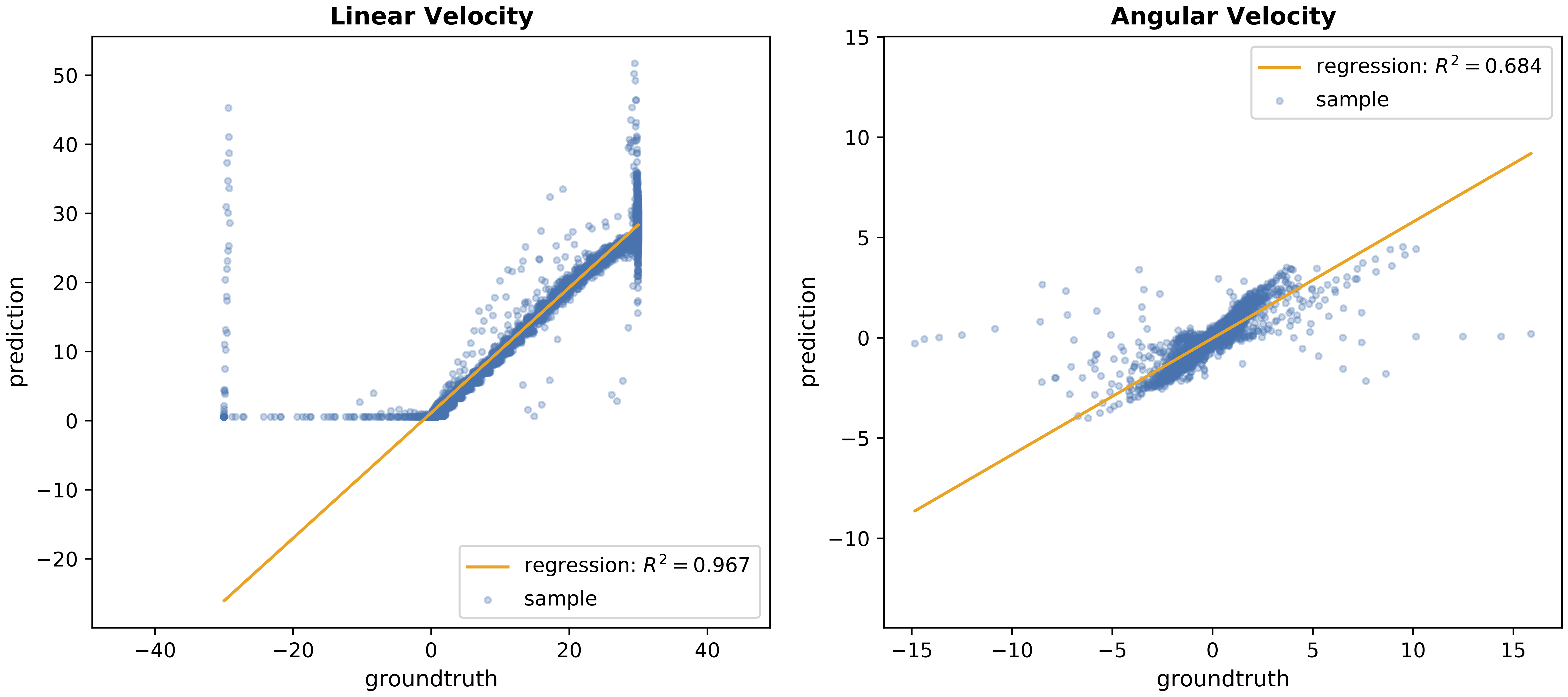}}
	\caption{$R^2$ regressor on the validation set.}
	\label{fig:regression-3}
\end{figure}

\begin{figure}[htbp]
	\centerline{\includegraphics[width=.8\columnwidth]{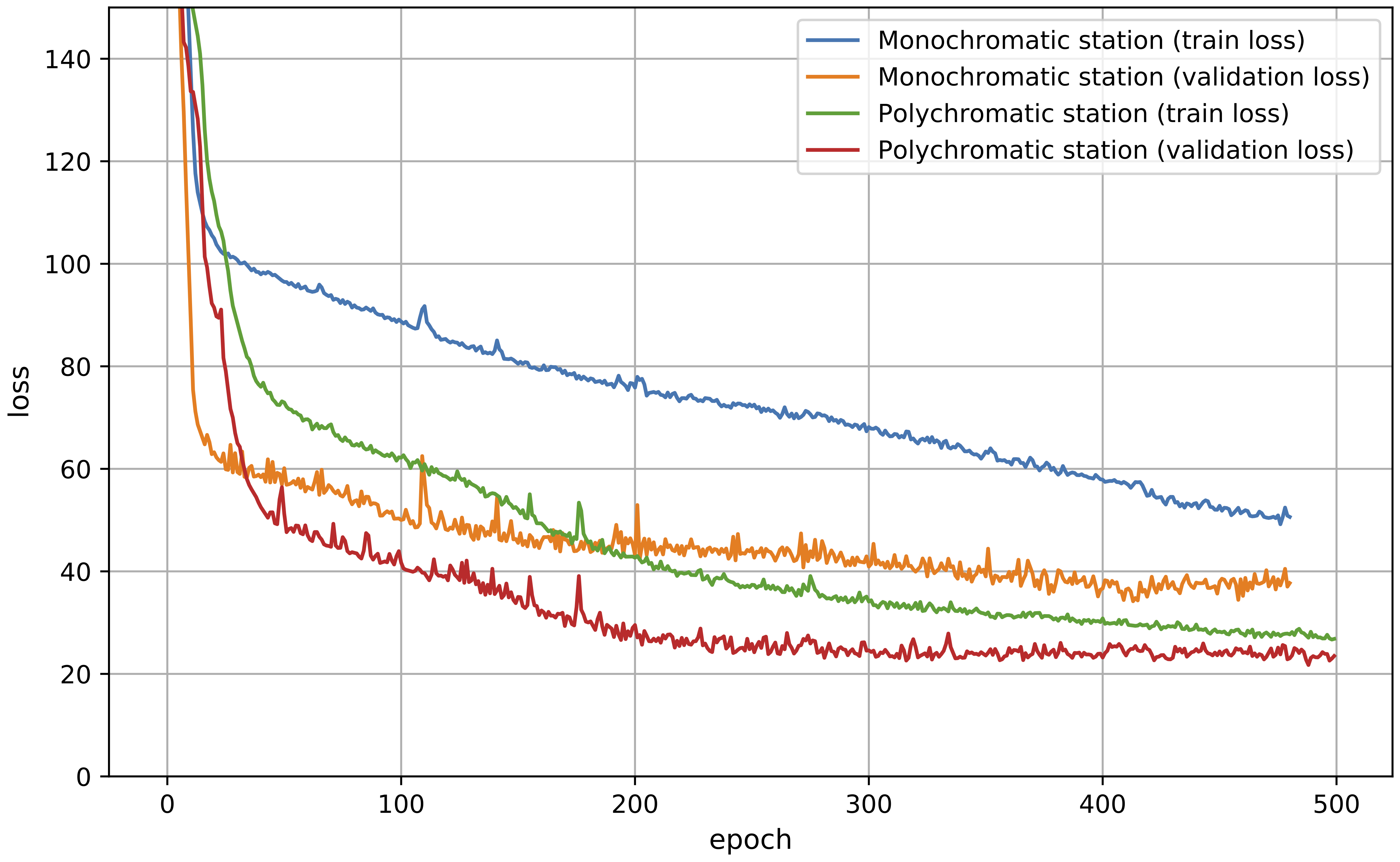}}
	\caption{Comparison of the losses among train and validation sets.}
	\label{fig:loss-3}
\end{figure}

The same network architecture and loss function of the first experiment are 
used to train the the model with a polychromatic object, and result in a 
significant improvement both in regression coefficient of the angular 
velocities, shown in Figure~\ref{fig:regression-3}, and in training and 
validation losses (in green and red), in Figure~\ref{fig:loss-3}.

\begin{figure}[htbp]
	\centerline{\includegraphics[width=\columnwidth]{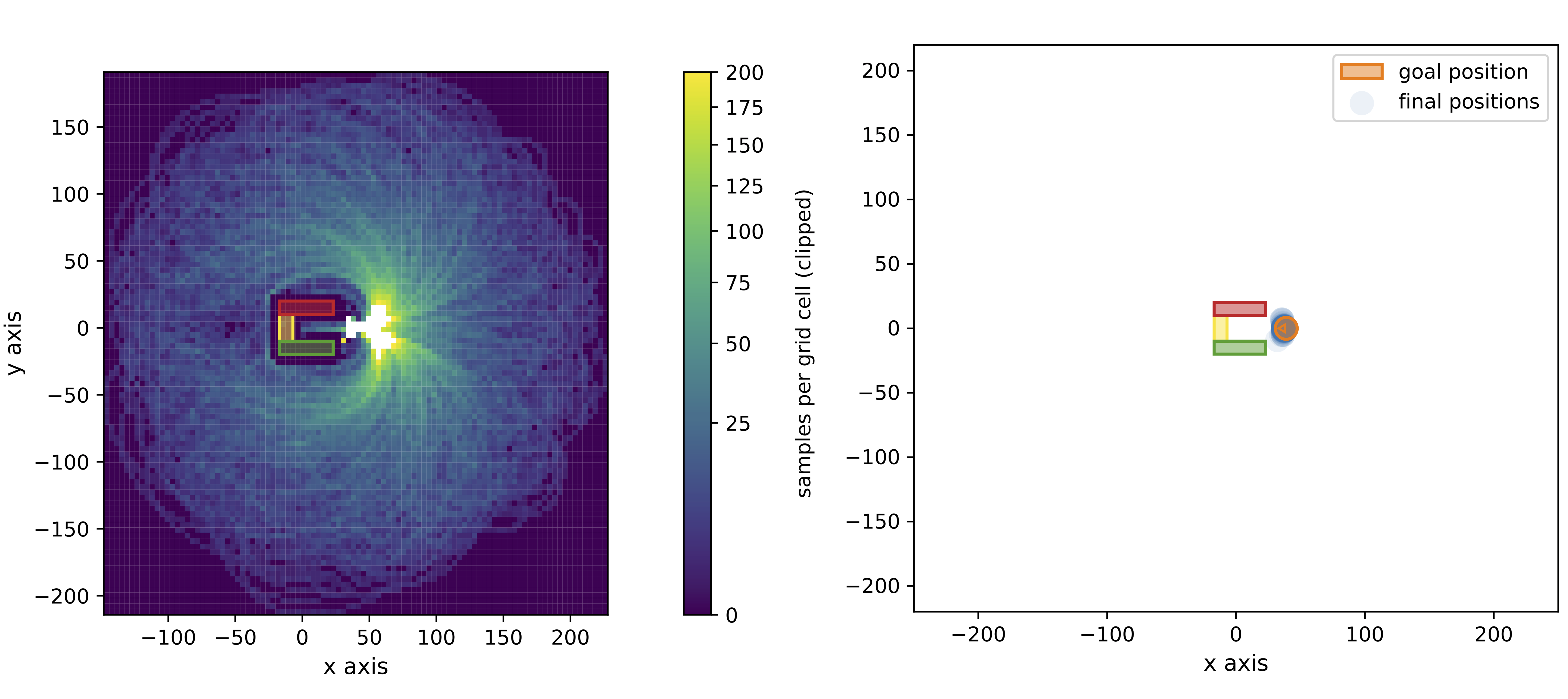}}
	\caption{Positions heatmap and final positions.}
	\label{fig:heatmap-final-positions}
\end{figure}

\begin{figure}[htbp]
	\centerline{\includegraphics[width=\columnwidth]{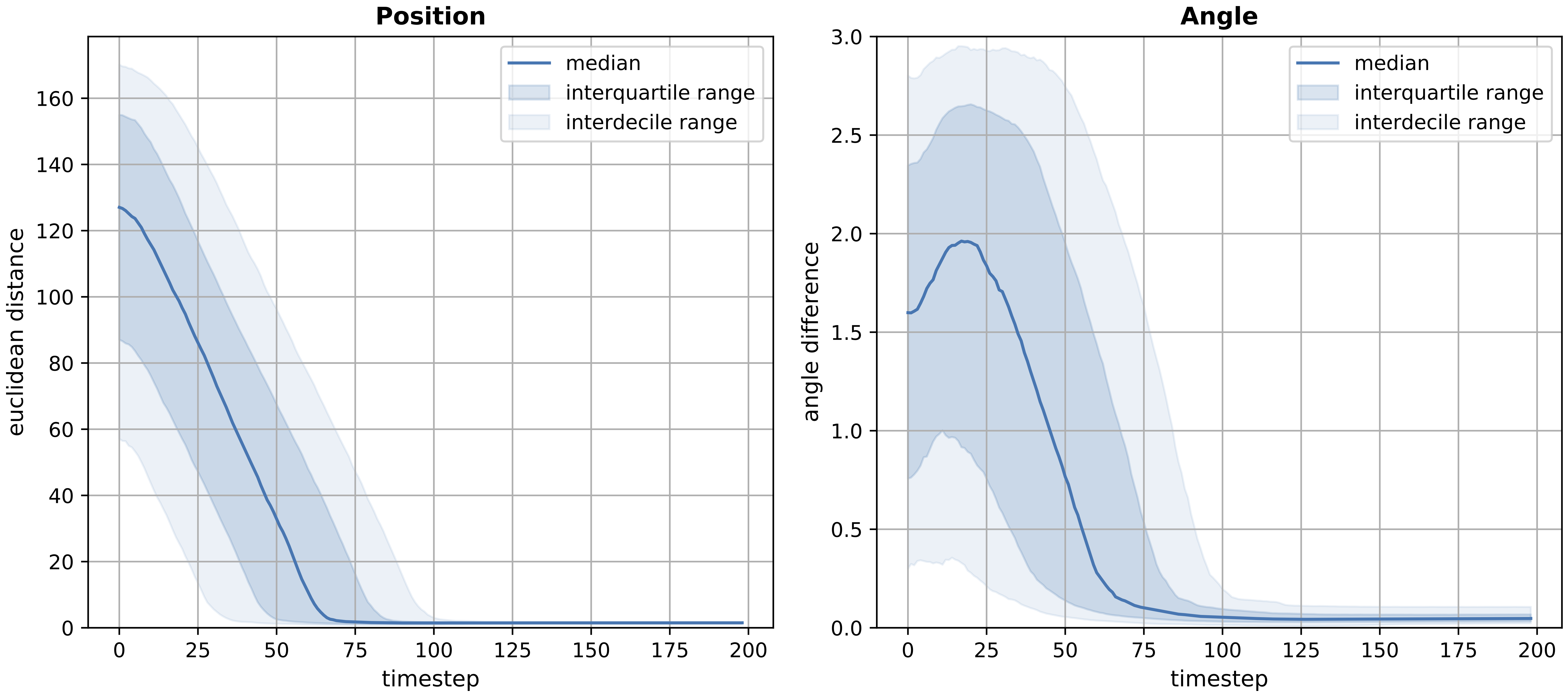}}
	\caption{Distance from goal over time.}
	\label{fig:distance-from-goal-learned3}
\end{figure}

Finally, Figure~\ref{fig:heatmap-final-positions} shows how the end positions 
are more tightly clustered over the goal than before. Moreover, the model is 
able to follow the optimal trajectories without rotating around the object.
Also in terms of convergence, the robot reaches the goal more precisely, as 
shown in Figure~\ref{fig:distance-from-goal-learned3}, and sometimes even 
faster than the omniscient controller.

\section{Task 2}
\label{sec:task2}
As an additional task, we experimented with arbitrary goal poses relative to a 
certain object. 

\begin{figure}[htbp]
	\centerline{\includegraphics[width=.8\columnwidth]{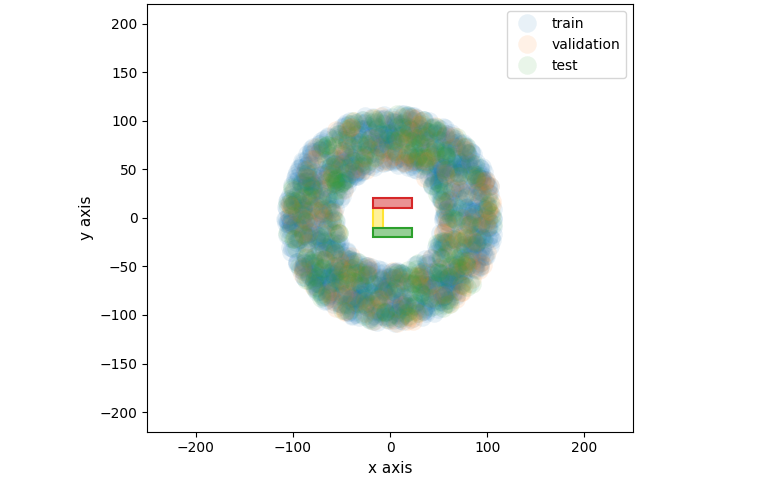}}
	\caption{Goal positions.}
	\label{fig:goal-positions}
\end{figure}

To this end, we first generated a dataset with the same omniscient controller 
as before, but with random goal poses located in a ring around the object, as 
shown in Figure~\ref{fig:goal-positions}.

Then, we implemented a neural network which receives the desired goal pose as 
input to the first fully-connected layer. The architecture used is shown in 
Table~\ref{tab: task 2}.

\begin{table}[htbp]
	\caption{Architecture of the Network for Task 2}
	\begin{center}
		\begin{tabular}{|c|c|c|c|c|}
			\hline
			\textbf{Layer}&\textbf{Channels} &\textbf{Kernel size} 
			&\textbf{Stride} &\textbf{Padding}\\
			\cline{1-5}
			conv1  &  4 $\rightarrow$ 	32 & 5 & 2 & 2, circular \\ \hline
			conv2  & 32 $\rightarrow$  	96 & 5 & 2 & 2, circular \\ \hline
			mpool1 & 					   & 3	& 3 & 1, circular \\ 
			\hline			
			conv3  & 96 $\rightarrow$  	96 & 5 & 1 & 2, circular \\ \hline
			fc1   &  1440 \textbf{+ 3} $\rightarrow$ 128 &  &  &  \\ \hline
			drop1 & \multicolumn{4}{c|}{dropout with p = 0.5} \\ \hline
			fc2   &  128 $\rightarrow$ 128 &  &  &  \\ \hline
			 drop2 & \multicolumn{4}{c|}{ dropout with p = 0.5} \\ \hline
			fc3 &  128 $\rightarrow$   2 &  &  &  \\ \hline
		\end{tabular}
		\label{tab: task 2}
	\end{center}
\end{table}

\begin{figure}[htbp]
	\centerline{\includegraphics[width=.8\columnwidth]{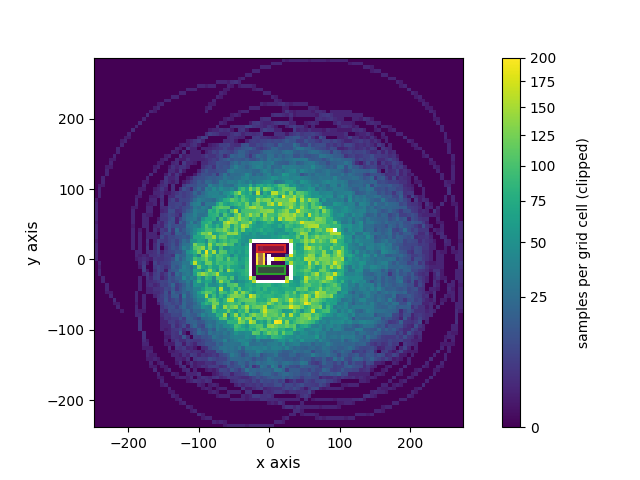}}
	\caption{Positions heatmap.}
	\label{fig:positions-heatmap-2}
\end{figure}

\begin{figure}[htbp]
	\centerline{\includegraphics[width=.8\columnwidth]{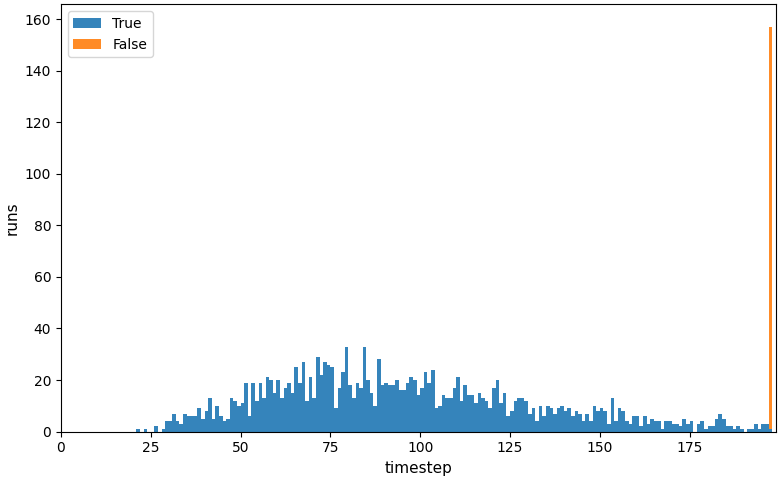}}
	\caption{Time to reach the goal.}
	\label{fig:goal-reached-2}
\end{figure}

From the plots shown in Figures~\ref{fig:positions-heatmap-2} and 
\ref{fig:goal-reached-2} we can see that the unchanged omniscient controller is 
not perfectly suited for this task: while it was possible to skip collision 
avoidance before, here it results in many runs getting stuck on the object.

\begin{figure}[htbp]
	\centerline{\includegraphics[width=\columnwidth]{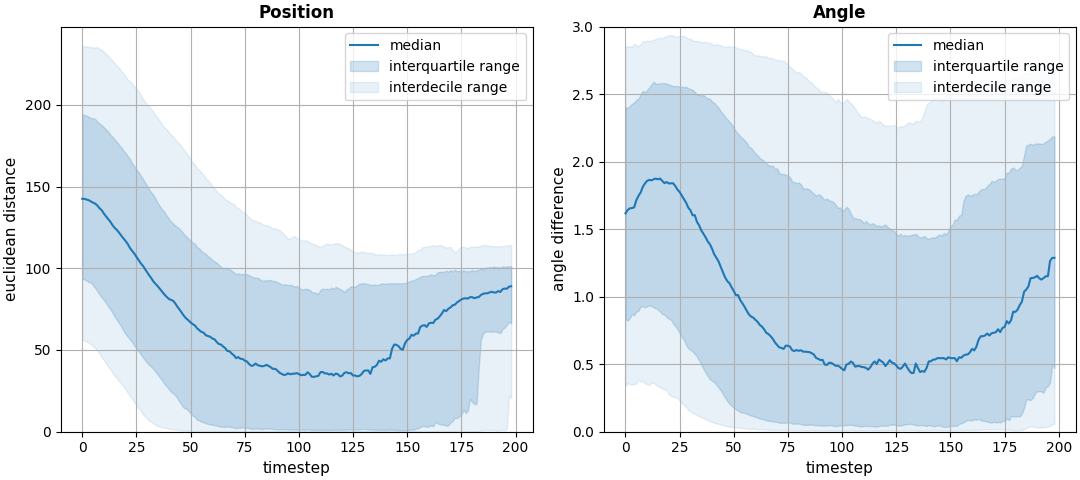}}
	\caption{Distance from goal.}
	\label{fig:distances-from-goal-2}
\end{figure}

The distances from goal, in Figure~\ref{fig:distances-from-goal-2}, confirm 
this issue, but in general the performance is good when there are no collisions.

\begin{figure}[htbp]
	\centerline{\includegraphics[width=\columnwidth]{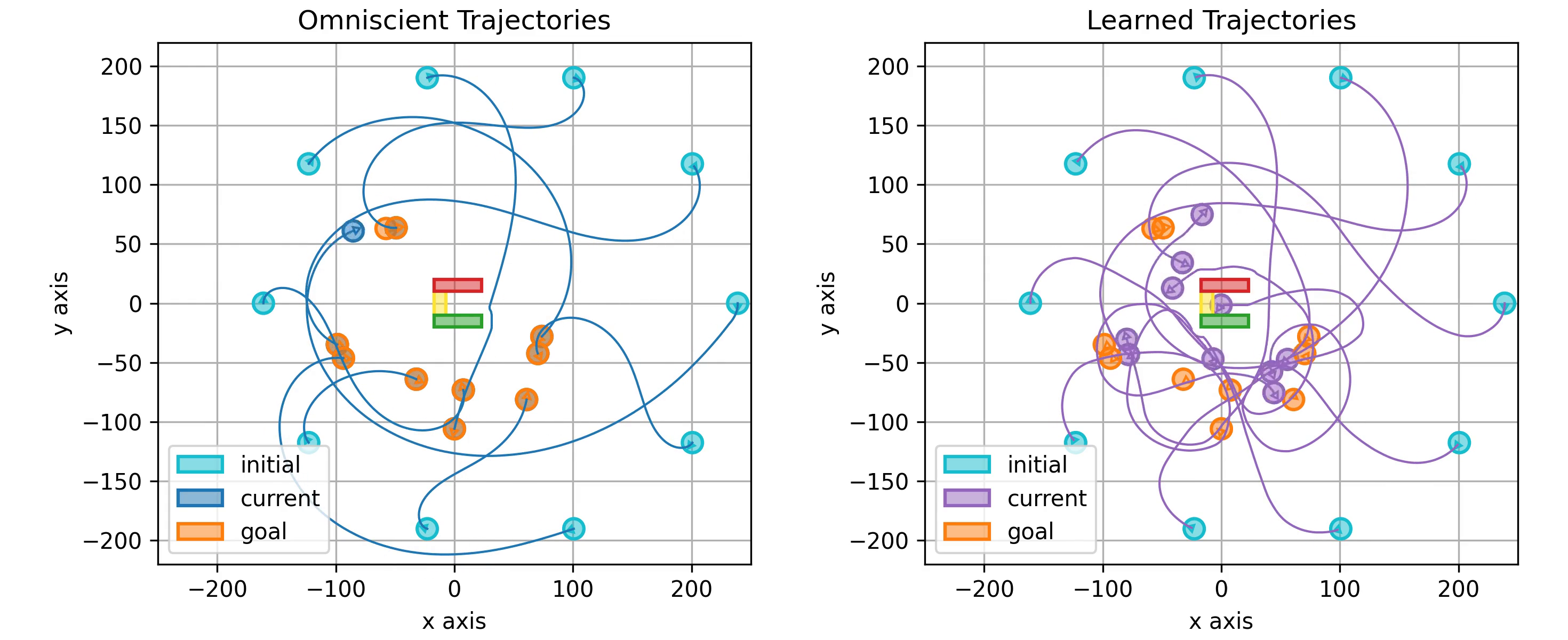}}
	\caption{Trajectories of the controller learned.}
	\label{fig:demo-circle-trajectories-2}
\end{figure}

This results in trajectories like those shown in 
Figure~\ref{fig:demo-circle-trajectories-2}.

\section{Future works}
In task 2, it appears that the performance of the neural network is quite good 
at moving in the general direction of the goal, less so at arriving with the 
correct orientation or stopping at the right time.

The first issue should be mitigated by a bigger architecture and more training 
data, since at the moment we are using only 2000 runs as before, which might 
not be enough given the higher complexity of the task.

The reason for the second problem is that the omniscient controller never 
overshoots the goal, so the network does not see this situation in training.
In Task 1, we solved the same issue with a special case: we had the omniscient 
controller move in reverse when it spawns inside the arms of the object. A 
similar solution could be applied here, extending this trick to arbitrary goal 
poses.

As a future work, we could try to see what happens to the current model when 
slightly changing the shape of the object. Moreover, we could also generalise 
the network by creating multiple polychromatic object with the faces coloured 
randomly. 

Another slightly more complex task could include adding obstacles that the 
robot should avoid, for example by orbiting around them.

In another spin-off, we can consider to add a second marXbot and learn to 
control the two robots with respect to each other (distributed control).


\bibliographystyle{IEEEtran}
\bibliography{IEEEabrv, biblio}

\end{document}